\documentclass[9pt]{article}

\usepackage{geometry}
\usepackage[utf8]{inputenc} 
\usepackage[T1]{fontenc}    
\usepackage{hyperref}       
\usepackage{url}            
\usepackage{booktabs}       
\usepackage{amsfonts}       
\usepackage{nicefrac}       
\usepackage{microtype}      
\usepackage{xcolor}         

\usepackage{enumitem}
\usepackage{caption,bm}
\usepackage{subfigure}
\usepackage{graphicx}
\usepackage{xspace,color}
\usepackage{diagbox,multirow} 
\usepackage{subfigure}
\usepackage{algorithm}
\usepackage{algorithmic}
\usepackage{amsmath,amssymb,bbm}

\usepackage[english]{babel}
\usepackage{amsthm}
\newtheorem{theorem}{Theorem}

\newtheorem{lemma}{Lemma}
\newtheorem{definition}{Definition}
\newtheorem{assumption}{Assumption}

\newcommand{\fullname}{differentiable scaffolding tree\ }
\newcommand{\mname}{\texttt{DST}\xspace }
\newcommand{\DST}{\texttt{DST}\xspace }
\newcommand{\bfx}{\mathbf{x}}

\newcommand{\bfA}{\mathbf{A}}

\newcommand{\bfE}{\mathbf{E}}

\newcommand{\bfb}{\mathbf{b}}
\newcommand{\bfB}{\mathbf{B}}
\newcommand{\bfH}{\mathbf{H}}

\newcommand{\bfN}{\mathbf{N}}
\newcommand{\bfS}{\mathbf{S}}
\newcommand{\bfw}{\mathbf{w}}

\newcommand{\bfU}{\mathbf{U}}

\newcommand{\bfV}{\mathbf{V}}

\newcommand{\RB}{\mathbb{R}}

\newcommand{\haty}{\widehat{y}}
\newcommand{\calT}{\mathcal{T}}
\newcommand{\calR}{\mathcal{R}}

\newcommand{\calO}{\mathcal{O}}
\newcommand{\calQ}{\mathcal{Q}}

\newcommand{\calV}{\mathcal{V}}
\newcommand{\calL}{\mathcal{L}}

\newcommand{\calW}{\mathcal{W}}
\newcommand{\calN}{\mathcal{N}}
\newcommand{\calS}{\mathcal{S}}

\newcommand{\calZ}{\mathcal{Z}}

\newcommand{\marked}{}

\title{Differentiable Scaffolding Tree for Molecular Optimization}

\author{Tianfan Fu$^{1*}$, Wenhao Gao$^{2*}$, Cao Xiao$^{3}$, Jacob Yasonik$^2$, Connor W. Coley$^{2}$ \& Jimeng Sun$^4$  \\
$^{1}$Georgia Institute of Technology, 
$^{2}$Massachusetts Institute of Technology\\
$^3$Amplitude. 
$^4$University of Illinois at Urbana-Champaign\\
$^{*}$Equal Contributions \\[2ex]%
Accepted by ICLR 2022 \\[2ex]
\texttt{tfu42@gatech.edu} \\
\texttt{\{whgao,jyasonik,ccoley\}@mit.edu}  \\
\texttt{\{danicaxiao,jimeng.sun\}@gmail.com} 
}

\date{\vspace{-5ex}}

\begin{document}

\maketitle

\begin{abstract}
The structural design of functional molecules, also called molecular optimization, is an essential chemical science and engineering task with important applications, such as drug discovery. 
Deep generative models and combinatorial optimization methods achieve initial success but still struggle with directly modeling discrete chemical structures and often heavily rely on brute-force enumeration. The challenge comes from the discrete and non-differentiable nature of molecule structures. 
To address this, we propose \fullname (\mname) that utilizes a learned knowledge network to convert discrete chemical structures to locally differentiable ones. 
\mname enables a gradient-based optimization on a chemical graph structure by back-propagating the derivatives from the target properties through a graph neural network (GNN). 
Our empirical studies show the gradient-based molecular optimizations are both effective and sample efficient. Furthermore, the learned graph parameters can also provide an explanation that helps domain experts understand the model output.
\end{abstract}


\section{Introduction}
\label{sec:intro}

The structural design of new functional molecules, also called molecular optimization, is the key to many scientific and engineering challenges, such as finding energy storage materials~\cite{hachmann2011harvard,janet2020accurate}, small molecule pharmaceutics~\cite{kuntz1992structure, zhavoronkov2019deep}, and environment-friendly material~\cite{zimmerman2020designing}. The objective is to identify novel molecular structures with desirable chemical or physical properties~\cite{gomez2018automatic,dai2018syntax,jin2018junction,You2018-xh,jin2019learning,shi2019graphaf,zhou2019optimization,jin2020multi,xie2021mars}. 
Recent advances in deep generative models (DGM) allow learning the distribution of molecules and optimizing the latent embedding vectors of molecules. Models in this category are exemplified by the variational autoencoder (VAE)~\cite{gomez2018automatic,dai2018syntax,jin2018junction,jin2020multi,guo2020property} and generative adversarial network (GAN)~\cite{de2018molgan}. 
On the other hand, because of the discrete and not explicitly combinatorial nature of the enormous chemical space, applying combinatorial optimization algorithms with some structure enumeration has been the predominant approach~\cite{You2018-xh,jensen2019graph,zhou2019optimization,nigam2019augmenting,xie2021mars}. 
Deep learning models have also been used to guide these combinatorial optimization algorithms. 
For example, \cite{You2018-xh,zhou2019optimization,jin2020multi,gottipati2020learning} tried to solve the problem with deep reinforcement learning; 
\cite{nigam2019augmenting} enhanced a genetic algorithm with a neural network as a discriminator; 
\cite{xie2021mars,fu2021mimosa} approached the problem with Markov Chain Monte Carlo (MCMC) to explore the target distribution guided by graph neural networks. 

Despite the initial success of these previous attempts, the following limitations remain: 
(1) deep generative models optimize the molecular structures in a learned latent space, which requires the latent space to be smooth and discriminative. Training such models needs carefully designed networks and well-distributed datasets.
(2) most combinatorial optimization algorithms, 
featured by evolutionary learning methods~\cite{nigam2019augmenting,jensen2019graph,xie2021mars,fu2021mimosa}, exhibit random-walk behavior, and leverage trial-and-error strategies to explore the discrete chemical space. 
The recent deep reinforcement learning methods~\cite{You2018-xh,zhou2019optimization,jin2020multi,gottipati2020learning} aim to remove random-walk search using a deep neural network to guide the searching. 
However, it is challenging to design the effective reward function into the objective~\cite{jin2020multi}. 
(3) Most existing methods require a great number of 
oracle calls (a property evaluator; see 
Def.~\ref{def:oracle}) to proceed with an efficient search. Realistic oracle functions, evaluating with either experiments or high-fidelity computational simulations, are usually expensive. Examples include using biological assays to determine the potency of drug candidates~\cite{wang2017pubchem}, or conducting electronic structure calculation to determine photoelectric properties~\cite{long2011electronic}.

Here we propose \fullname (\mname) to address these challenges, where we define a differentiable scaffolding tree for molecular structure and utilize a trained GNN to obtain the local derivative that enables continuous optimization. The main contributions are summarized as follows: 
\begin{itemize}[leftmargin=5.5mm]
\item We propose the differentiable scaffolding tree to define a local derivative of a chemical graph. 
This concept enables a gradient-based optimization of a discrete graph structure. 


\item We present a general molecular optimization strategy utilizing the local derivative defined by the differentiable scaffolding tree. This strategy leverages the property landscape's geometric structure and suppresses the random-walk behavior, exploring the chemical space more efficiently. 
We also incorporate a determinantal point process (DPP) based selection strategy to enhance the diversity of generated molecules.  

\item We demonstrate encouraging preliminary results on \textit{de novo} molecular optimization with multiple computational objective functions. The local derivative shows consistency with chemical intuition, providing interpretability of the chemical structure-property relationship. Our method also requires less oracle calls, maintaining good performance in limited oracle settings. 
\end{itemize}

\section{Related Work}
\label{sec:related}


Existing molecular optimization methods can mainly be categorized as deep generative models and combinatorial optimization methods. 

\noindent\textbf{Deep generative models} model a distribution of general molecular structure with a deep network model so that one can generate molecules by sampling from the learned distribution. Typical algorithms include variational autoencoder (VAE), generative adversarial network (GAN), energy-based models, flow-based model
~\cite{gomez2018automatic, jin2018junction, de2018molgan, segler2018generating,  jin2019learning,honda2019graph,madhawa2019graphnvp,shi2019graphaf, jin2020multi, kotsias2020direct,liu2021graphebm,fu2021probabilistic,bagal2021liggpt}. 
Also, DGMs can leverage Bayesian optimization in latent spaces to optimize latent vectors and reconstruct to obtain the optimized molecules~\cite{jin2018junction}. 
However, such approaches usually require a smooth and discriminative latent space and thus an elaborate network architecture design and well-distributed data set. 
Also, as they learn the reference data distribution, their ability to explore diverse chemical space is relatively limited, evidenced by the recent molecular optimization benchmarks~\cite{brown2019guacamol, huang2021therapeutics}.


\noindent\textbf{Combinatorial optimization methods} mainly include deep reinforcement learning (DRL)~\cite{You2018-xh,zhou2019optimization,jin2020multi,gottipati2020learning} and evolutionary learning methods~\cite{nigam2019augmenting,jensen2019graph,xie2021mars,fu2021mimosa}. They both formulate molecule optimization as a discrete optimization task. 
Specifically, they modify molecule substructures (or tokens in a string representation~\cite{weininger1988smiles}) locally, with an oracle score or a policy/value network to tell if they keep it or not. 
Due to the discrete nature of the formulation, most of them conduct an undirected search (random-walk behavior), while some recent ones like reinforcement learning try to guide the searching with a deep neural network, aiming to rid the random-walk nature. 
However, it is challenging to incorporate the learning objective target into the guided search. 
Those algorithms still require massive numbers of oracle calls, which is computationally inefficient during the inference time~\cite{korovina2020chembo}.
Our method, \mname, falls into this category, explicitly leverages the objective function landscape and conducts an efficient goal-oriented search. Instead of operating on molecular substructure or tokens, we define the search space as a set of binary and multinomial variables to indicate the existence and identity of nodes respectively, and make it locally differentiable with a learned GNN as a surrogate of the oracle. This problem formulation can find its root in conventional computer-aided molecular design algorithms with branch-and-bound algorithms as solutions~\cite{sinha1999environmentally, sahinidis2000applications}.




\section{Method}
\label{sec:method}

\begin{figure}[t]
\centering
\subfigure{\includegraphics[width=\linewidth]{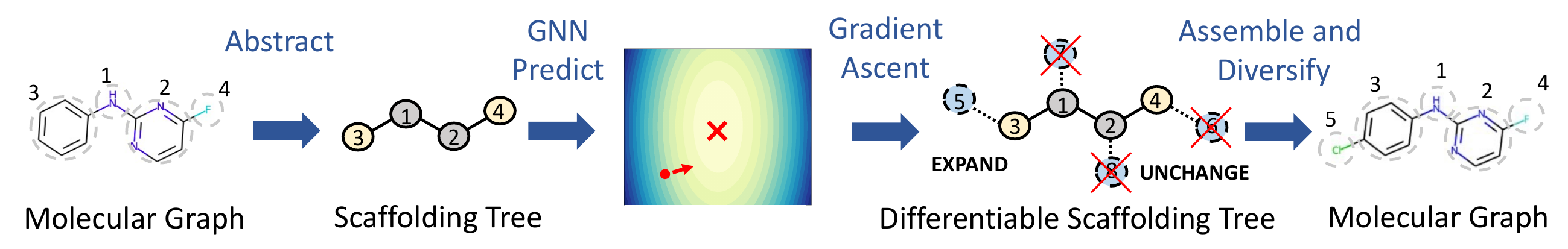}}
\caption{
Illustration of the overall approach: 
During inference, we construct the corresponding scaffolding tree and differentiable scaffolding tree (\DST) for each molecule.
We optimize each \DST along its gradient back-propagated from the GNN and sample scaffolding trees from the optimized \DST. After that, we assemble trees into molecules and diversify them for the next iteration. 
}
\label{fig:framework}
\end{figure}

We first introduce the formulation of molecular optimization and \textit{differentiable scaffolding tree} (\mname) in Section~\ref{sec:formulation_notation}, illustrate the pipeline in Figure~\ref{fig:framework}, then describe the key steps following the order:
\begin{itemize}[leftmargin=*]
\item {\bf Oracle GNN construction:} We leverage GNNs to imitate property oracles, which are targets of molecular optimization (Section~\ref{sec:gnn}). {Oracle GNN is trained once and for all. The training is separately from optimizing \mname below. }
\item {\bf Optimizing differentiable scaffolding tree:}  We formulate the discrete molecule optimization into \textit{a locally differentiable} problem with a differentiable scaffolding tree (\mname). Then a \mname can be optimized by the gradient back-propagated from oracle GNN (Section~\ref{sec:differentiable}). 
\item {\bf Molecule Diversification} After that, we describe how we design a \textit{determinantal point process (DPP)} based method to output diverse molecules for iterative learning (Section~\ref{sec:dpp}). 
\end{itemize}

\subsection{Problem Formulation and Notations}
\label{sec:formulation_notation}

\noindent\textbf{3.1.1\ \  Molecular optimization problem}\ \ 
Oracles are the objective functions for molecular optimization problems, e.g., QED quantifying a molecule's drug-likeness~\cite{bickerton2012quantifying}. 
\begin{definition}[Oracle $\calO$]
\label{def:oracle}
Oracle $\calO$ is a black-box function that evaluates certain chemical or biological properties of a molecule $X$ and returns the ground truth property $\calO(X)$.   
\end{definition}
In realistic discovery settings, the oracle acquisition cost is usually not negligible. 
Suppose we want to optimize $P$ molecular properties specified by oracle $\calO_1, \cdots, \calO_{P}$, we can formulate a multi-objective molecule optimization problem through scalarization as represented in Eq.~\eqref{eqn:objective}, 
\begin{equation}
\label{eqn:objective}
\begin{aligned}
{\arg\max}_{X \in \calQ} \ \  F(X; \calO_1, \calO_2, \cdots, \calO_{P}) = f(\calO_1(X), \cdots, \calO_{P}(X)), 
\end{aligned}
\end{equation}
where $X$ is a molecule, $\calQ$ denotes the set of valid molecules; $f$ is the composite objective combining all the oracle scores, e.g., the mean value of $P$ oracle scores.

\noindent\textbf{3.1.2\ \  Scaffolding Tree}\ \ 
The basic mathematical description of a molecule is molecular graph, which contains atoms as nodes and chemical bonds as edges. 
However, molecular graphs are not easy to generate explicitly as graphs due to the presence of rings, relatively large size, and chemical validity constraints. 
For ease of computation, we convert a molecular graph to a scaffolding tree as a higher-level representation, a tree of substructures, following~\cite{jin2018junction,jin2019learning}. 
\begin{definition}[\textbf{Substructure}]
Substructures can be either an atom or a single ring. The substructure set is denoted $\calS$ (vocabulary set), which covers frequent atoms and single rings in drug-like molecules.
\end{definition}
\begin{definition}[\textbf{Scaffolding Tree $\calT$}]
\label{def:tree}
A scaffolding tree, $\calT_X$, is a spanning tree whose nodes are substructures. It is  higher-level representation of molecular graph $X$. 
\end{definition}

$\calT_X$ is represented by (i) node indicator matrix, (ii) adjacency matrix, and (iii) node weight vector. 
We distinguish leaf and non-leaf nodes in $\calT_X$.
Among the $K$ \footnote{{$K$ depends on molecular graph. During optimization (Section~\ref{sec:differentiable} and~\ref{sec:dpp}), after molecular structure changes, $K$ is updated.}} nodes in $\calT_X$, there are $K_{\text{leaf}}$ leaf nodes (nodes connecting to only one edge) and $K - K_{\text{leaf}}$ non-leaf nodes (otherwise). The sets of leaf nodes and non-leaf nodes are denoted $\calV_{\text{leaf}}$ and $\calV_{\text{nonleaf}}$ correspondingly. 
\begin{definition}
\label{def:node_index_matrix}
\textbf{Node indicator matrix} $\bfN$ is decomposed as $\bfN = \begin{pmatrix} \bfN_{\text{nonleaf}} \\ \bfN_{\text{leaf}} \ \ \ \ \ \end{pmatrix} \in \{0,1\}^{K \times |\calS|}$,  
where $\bfN_{\text{nonleaf}} \in \{0,1\}^{(K-K_{\text{leaf}}) \times |\calS|}$ corresponds to non-leaf nodes while $\bfN_{\text{leaf}} \in \{0,1\}^{ K_{\text{leaf}} \times |\calS|}$ corresponds to leaf nodes. 
Each row of $\bfN$ is a one-hot vector, indicating which substructure the node belongs to. 
\end{definition}

\begin{definition}
\label{def:adj}
\textbf{Adjacency matrix} is denoted $\bfA \in \{0,1\}^{K\times K}$. $A_{ij} = 1$ indicates the $i$-th node and $j$-th node are connected while 0 indicates unconnected. 
\end{definition}

\begin{definition}
\label{def:node_weight}
\textbf{Node weight vector}, $\bfw = [1,\cdots,1]^\top \in \RB^K$, indicates the $K$ nodes in scaffolding tree are equally weighted. 
\end{definition}


\begin{figure}[t]
\centering
\includegraphics[width=\linewidth]{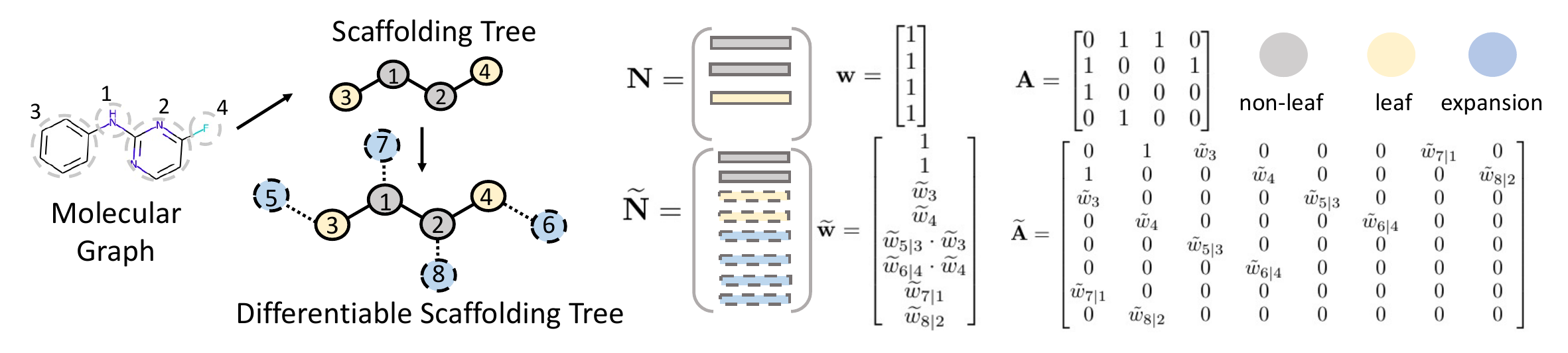}
\caption{Example of differentiable scaffolding tree. 
We show non-leaf nodes (grey), leaf nodes (yellow), expansion nodes (blue). 
The dashed nodes and edges are learnable, corresponding to nodes' identity and existence, respectively. 
$\widetilde{\bfw}$ and $\widetilde{\bfA}$ share the learnable parameters $\{\widehat{\bfw}_3, \widehat{\bfw}_4, \widehat{\bfw}_{5|3}, \widehat{\bfw}_{6|4}, \widehat{\bfw}_{7|1}, \widehat{\bfw}_{8|2} \}$.
}
\label{fig:adj}
\end{figure}

\noindent\textbf{3.1.3\ \  Differentiable scaffolding tree}\ \
Similar to a scaffolding tree, a differentiable scaffolding tree (DST) also contains (i) node indicator matrix, (ii) adjacency matrix, and (iii) node weight vector, but with additional expansion nodes. Specifically, while inheriting leaf node set $\calV_{\text{leaf}}$ and non-leaf node set $\calV_{\text{nonleaf}}$ from the original scaffolding tree, we add expansion nodes and form \underline{\textit{expansion node set}}, $\calV_{\text{expand}} = \{u_v | v \in \calV_{\text{leaf}}\cup \calV_{\text{nonleaf}} \}, |\calV_{\text{expand}}|=K_{\text{expand}}=K$, where $u_{v}$ is connected to $v$ in the original scaffolding tree. 
We also define \underline{\textit{differentiable edge set}}, $\Lambda = \{(v, v')\ |\ v\in \calV_{\text{leaf}} \ \text{OR}\ v'\in \calV_{\text{expand}};\ v,v'\ \text{are connected} \}$ to incorporate all the edges involving leaf-nonleaf node and leaf/nonleaf-expansion node connections. To make it locally differentiable, we modify the tree parameters from two aspects: (A) {\it node identity} and (B) {\it node existence}. 
Figure~\ref{fig:adj} shows an example to illustrate DST. 

(A) We enable optimization on {\it node identity} by allowing the corresponding node indicator matrix learnable:
\begin{definition}
\label{def:dif_node_index_matrix}
\textbf{Differentiable node indicator matrix} $\widetilde{\bfN}$ takes the form: 
\begin{equation}
\label{eqn:differentiable_node_index_matrix}
\widetilde{\bfN} = 
\begin{pmatrix} \widetilde{\bfN}_{\text{nonleaf}} \\ \widetilde{\bfN}_{\text{leaf}} \ \ \ \ \  \\ \widetilde{\bfN}_{\text{expand}} \end{pmatrix} \in \RB_{+}^{(K+K_{\text{expand}}) \times |\calS|}, \ \ \ \sum_{j=1}^{|\calS|} \widetilde{\bfN}_{ij} = 1, \ \ K=K_{\text{expand}}.
\end{equation} 
$\widetilde{\bfN}_{\text{nonleaf}} = {\bfN}_{\text{nonleaf}} \in \{0,1\}^{(K-K_{\text{leaf}}) \times |\calS|}$ are fixed, equal to the part in the original scaffolding tree, each row is a one-hot vector, indicating that we fix all the non-leaf nodes. 
In contrast, both $\widetilde{\bfN}_{\text{expand}}$ and $\widetilde{\bfN}_{\text{leaf}}$ are learnable, we use softmax activation to implicitly encode the constraint $\sum_j\widetilde{\bfN}_{ij} = 1$
i.e., $\widetilde{\bfN}_{ij} = \frac{\exp(\widehat{\bfN_{ij}})}{\sum_{j'=1}^{|\calS|} \exp(\widehat{\bfN_{i,j'}})}$, $\widehat{\bfN}$ are the parameters to learn. This constraint guarantee that each row of $\widetilde{\bfN}$ is a valid substructures' distribution. 
\end{definition}



(B) We enable optimization on {\it node existence} by assigning learnable weights for the leaf and expansion nodes, construct adjacency matrix and node weight vector based on those values: 
\begin{definition}
\label{def:dif_adj}
\textbf{Differentiable adjacency matrix}
$\widetilde{\bfA} \in \RB^{(K+K_{\text{expand}})\times (K+K_{\text{expand}})}$ takes the form:
\begin{equation}
\label{eqn:diff_adj}
\widetilde{\bfA}_{ij} = \widetilde{\bfA}_{ji} = 
\left\{
\begin{array}{ll}
 1/0,              & (i,j) \notin \Lambda\  \ \text{0:disconnected,\ \ 1:connected} \\
\sigma(\widehat{\bfw}_{i}),        &  (i,j) \in \Lambda, i\in \calV_{\text{leaf}},\ j\in \calV_{\text{nonleaf}} \\
\sigma(\widehat{\bfw}_{i|j}),        &  (i,j) \in \Lambda, i\in \calV_{\text{expand}},\ j\in \calV_{\text{leaf}} \cup \calV_{\text{nonleaf}} \\
\end{array}
\right. 
\end{equation}
where $\Lambda$ is the differentiable edge set defined above,  
Sigmoid function $\sigma(\cdot)$ imposes the constraint $0\leq\widetilde{\bfA}_{ij}\leq 1$. 
$\widehat{\bfw} \in \RB^{K_{\text{leaf}} + K_{\text{expand}}}$ are the parameters, each leaf node and expansion node has one learnable parameter. 
For connected $i$ and $j$, when $i\in \calV_{\text{leaf}},\ j\in \calV_{\text{nonleaf}}$, $\widetilde{\bfA}_{ij} = \sigma(\widehat{\bfw}_{i})$ measures the existence probability of leaf node $i$; when $i\in \calV_{\text{expand}}, j\in \calV_{\text{leaf}} \cup \calV_{\text{nonleaf}}$, $\widetilde{\bfA}_{ij} = \sigma(\widehat{\bfw}_{i|j})$ measures the conditional probability of the existence of expand node $i$ given the original node $j$. 
When $j$ is a leaf node, it naturally embeds the inheritance relationship between the leaf node and the corresponding expansion node. 
\end{definition}

\begin{definition}
\label{def:dif_node_weight}
\textbf{Differentiable node weight vector}
$\widetilde{\bfw} \in \RB^{K+K_{\text{expand}}}$ takes the form: 
\begin{equation}
\label{eqn:diff_node_weight}
\widetilde{\bfw}_{i}=
\left\{
\begin{array}{ll}
1,  & i\in \calV_{\text{nonleaf}} \\
\sigma(\widehat{\bfw}_i),  &  i\in \calV_{\text{leaf}}\\
\sigma(\widehat{\bfw}_{i|j}) \sigma(\widehat{\bfw}_j),  &  i\in \calV_{\text{expand}}, j\in \calV_{\text{leaf}},\ \ \ \ (i,j)\in \Lambda, \\
\sigma(\widehat{\bfw}_{i|j}) \tilde{\bfw}_j = \sigma(\widehat{\bfw}_{i|j}),  &  i\in \calV_{\text{expand}}, j\in \calV_{\text{nonleaf}},\ \ \ \ (i,j)\in \Lambda, \\
\end{array}
\right. 
\end{equation}
where all the weights range from 0 to 1. The weight of expansion node connecting to leaf node relies on the weight of corresponding leaf node. $\widetilde{\bfw}$ and $\widetilde{\bfA}$ (Def.~\ref{def:dif_adj}) shares the learnable parameter $\widehat{\bfw}$. 
Figure~\ref{fig:adj} shows an example to illustrate DST. 
\end{definition}

\subsection{Training Oracle Graph Neural Network}
\label{sec:gnn}

This section constructs a differentiable surrogate model to capture the knowledge from any oracle function. We choose graph neural network architecture for its state-of-the-art performance in modeling structure-property relationships. In particular, we imitate the objective function $F$ with GNN:
\begin{equation}
\label{eqn:gnn}
\begin{aligned}
\haty = \text{GNN}(X; \Theta) \approx F(X; \calO_1, \calO_2, \cdots, \calO_P) = y, 
\end{aligned}
\end{equation}
where $\Theta$ represents the GNN's parameters. 
Concretely, we use a graph convolutional network (GCN)~\cite{kipf2016semi}. Other GNN variants, such as Graph Attention Network (GAT)~\cite{velivckovic2017graph}, Graph Isomorphism Network (GIN)~\cite{xu2018powerful}, can also be used in our setting.  
The initial node embeddings $\bfH^{(0)} = \bfN \bfE \in \RB^{K\times d}$ stacks basic embeddings of all the nodes in the scaffolding tree, $d$ is the GCN hidden dimension, $\bfN$ is the node indicator matrix (Def.~\ref{def:node_index_matrix}).  $\bfE \in \RB^{|\calS|\times d}$ is the embedding matrix of all the substructures in vocabulary set $\calS$, and is randomly initialized. 
The updating rule of GCN for the $l$-th layer is 
\begin{equation}
\label{eqn:gcn}
\begin{aligned}
& \bfH^{(l)} = \text{RELU}\big( \bfB^{(l)} +  \bfA(\bfH^{(l-1)}\bfU^{(l)}) \big), \ \ \  l=1,\cdots,L, 
\end{aligned}
\end{equation}
where $L$ is GCN's depth, $\bfA$ is the adjacency matrix (Def.~\ref{def:adj}), $\bfH^{(l)}\in\RB^{K\times d}$ is the nodes' embedding of layer $l$, $\bfB^{(l)} = [{\bfb^{(l)}, \bfb^{(l)}, \cdots, \bfb^{(l)}}]^\top \in \RB^{K\times d}$ and $\bfU^{(l)} \in \RB^{d\times d}$ are bias and weight parameters of layer $l$, respectively.  

We generalize the GNN from a discrete scaffolding tree to a differentiable one.
Based on learnable weights for each node, we leverage the weighted average as the readout function of the last layer's ($L$-th) node embeddings, followed by multilayer perceptron (MLP) to yield the prediction $\haty$, i.e., $\haty = \text{MLP}\big(\frac{1}{\sum_{k=1}^{K} w_k} \sum_{k=1}^{K} w_k H_{k}^{(L)} \big)$, in discrete scaffolding tree, weights for all the nodes are equal to 1, $H^{(L)}_k$ is the $k$-th row of $H^{(L)}$. 
In sum, the prediction can be written as 
\begin{equation}
\label{eqn:gnn_prediction}
\haty = \text{GNN}(X; \Theta) = \text{GNN}(\calT_X = \{\bfN, \bfA, \bfw\}; \Theta), \ \ \ X \in \calQ
\end{equation}
where $\Theta = \{\bfE\}\cup \{\bfB^{(l)}, \bfU^{(l)}\}_{l=1}^{L}$ are the GNN's parameters. 
We train the GNN by minimizing the discrepancy between GNN prediction $\haty$ and the ground truth $y$. 
\begin{equation}
\label{eqn:gnn_objective}
\begin{aligned}
\Theta_{*} & = \underset{\Theta}{\arg\min} \sum_{(X,y)\in \mathcal{D}}^{} \calL\big(y = F(X; \calO_1, \calO_2, \cdots, \calO_P), \ \ \widehat{y} = \text{GNN}(X;\Theta) \big), \\ 
\end{aligned}
\end{equation}
where $\calL$ is loss function, e.g., mean squared error; $\mathcal{D}$ is the training set. 
After training, we have GNN parameterized by $\Theta_{*}$ to approximate the black-box objective function $F$ (Eq.~\ref{eqn:objective}). 
Worth to mention that Oracle GNN is trained once and for all. The training is separately from optimizing \mname below.


\subsection{Optimizing Differentiable Scaffolding Tree}
\label{sec:differentiable}

\textbf{Overview\ \ } 
With a little abuse of notations, 
via introducing \mname, we approximate molecule optimization as a \textit{locally differentiable} problem  
\begin{equation}
\begin{aligned}
& \underbrace{X = {\arg\max}_{X\in \calQ}\ F(X^{})}_{\text{(I) structured combinatorial optimization}}
\ \ \   {\approx} \ \ \ 
\underbrace{X^{(t+1)} = {\arg\max}_{X\in \calN(X^{(t)})}\ F(X^{}) }_{\text{(II) iterative local discrete search}}  \\  
\ \ \  &  {\approx} \ \ \ 
\underbrace{\calT_{X^{(t+1)}}  = \underset{ X \in \calN(X^{(t)}) }{\arg\max} \ \text{GNN}( \calT_X= \{ \widetilde{\bfN}_{X^{(t)}}, \widetilde{\bfA}_{X^{(t)}}, \widetilde{\bfw}_{X^{(t)}}\}; \Theta_{*})}_{\text{(III) local differentiable optimization}}, 
\end{aligned}
\end{equation}
where $X^{(t)}$ is the molecule at $t$-th iteration, $\calN(X^{(t)}) \subseteq \calQ$ is the neighborhood set of $X^{(t)}$ (Def.~\ref{def:neighbor}). 
Next, we explain the intuition behind these approximation steps. Molecular optimization is generally a discrete optimization task, which is prohibitively expensive due to exhaustive search. 
The first approximation is to formulate the problem as an iterative local discrete search via introducing a neighborhood molecule set $\calN(X^{(t)})$. Second, to enable differentiable learning, we use GNN to imitate black-box objective $F$ (Section~\ref{sec:gnn}) and further reformulated it into \textit{a local differentiable optimization} problem. Then we can optimize \mname ($\calT_X= \{ \widetilde{\bfN}_{X^{(t)}}, \widetilde{\bfA}_{X^{(t)}}, \widetilde{\bfw}_{X^{(t)}}\}$) in a continuous domain for $\calN(X^{(t)})$ using gradient-based optimization method.



\noindent\textbf{3.3.1\ \  Local Editing Operations}\ \ 
For a leaf node $v$ in the scaffolding tree, we can perform three editing operations, (1) \textbf{SHRINK}: delete node $v$; (2) \textbf{REPLACE}: replace a new substructure over $v$; (3) \textbf{EXPAND}: add a new node $u_v$ that connects to node $v$. 
For a nonleaf node $v$, we support (1) \textbf{EXPAND}: add a new node $u_v$ connecting to $v$; (2) \textbf{do nothing}. 
If we EXPAND and REPLACE, the new substructures are sampled from the vocabulary $\calS$. 
We define a molecule neighborhood set as below: 
\begin{definition}[Neighborhood set] 
\label{def:neighbor}
Neighborhood set of molecule $X$, denoted $\calN(X)$, 
is the set of all the possible molecules obtained by imposing one local editing operation to scaffolding tree $\calT_X$ and assembling the edited trees into molecules. 
\end{definition}



\noindent\textbf{3.3.2\ \  Optimizing DST}\ \ 
Then within the domain of neighborhood molecule set $\calN(X)$, the objective function can be represented as a differentiable function of X's DST ($\widetilde{\bfN}_X, \widetilde{\bfA}_X, \widetilde{\bfw}_X$). We address the following optimization problem to get the best scaffolding tree within $\calN(X)$, 
\begin{equation}
\label{eqn:optimize}
\widetilde{\bfN}_*, \widetilde{\bfA}_*, \widetilde{\bfw}_* = {\arg\max}_{ \{ \widetilde{\bfN}_X, \widetilde{\bfA}_X, \widetilde{\bfw}_X \} } \ \ \text{GNN}(\{ \widetilde{\bfN}_X, \widetilde{\bfA}_X, \widetilde{\bfw}_X \}; \Theta_{*}), 
\end{equation}
where the GNN parameters $\Theta_{*}$ (Eq.~\eqref{eqn:gnn_objective}) are fixed. 
Comparing with Eq.~\eqref{eqn:gnn_prediction}, it is differentiable with regard to $\{ \widetilde{\bfN}, \widetilde{\bfA}, \widetilde{\bfw} \}$ for all molecules in the neighborhood set $\calN(X)$. Therefore, we can optimize the \mname using gradient-based optimization method, e.g., an Adam optimizer~\cite{kingma2014adam}.

\noindent\textbf{3.3.3\ \ Sampling from DST}\ \ 
Then we sample the new scaffolding tree from the optimized \mname. 
Concretely, (i) for each leaf node $v \in \calV_{\text{leaf}}$ and the corresponding expansion node $u_v \in \calV_{\text{expand}}$, we select one of the following step with probabilities (w.p.) as follows, 
\begin{equation}
\label{eqn:sample}
\begin{aligned}
& \calT \sim \ \text{DST-Sampler}(\widetilde{\bfN}_*, \widetilde{\bfA}_*, \widetilde{\bfw}_*) \\
= & \left\{
\begin{array}{ll}
\text{1. SHRINK: delete leaf node\ }v, &  \text{w.p. $1-(\widetilde{\bfw}_{*})_{v}$}, \\
\text{2. EXPAND: add $u_v$, select substructure at $u_v$ based on $(\widetilde{\bfN}_*)_{u_v}$}, &    \text{w.p. $(\widetilde{\bfw}_{*})_v (\widetilde{\bfw}_{*})_{u_v|v}$}, \\
\text{3. REPLACE: select substructure at $v$ based on $(\widetilde{\bfN}_*)_{u}$},  &  \text{w.p. $(\widetilde{\bfw}_{*})_v (1-(\widetilde{\bfw}_{*})_{u_v|v})$}. \\
\end{array}
\right.
\end{aligned}
\end{equation}
(ii) For each nonleaf node $v$, we expand a new node with probability $(\widetilde{\bfw}_{*})_{u_v|v}$. If expanding, we select substructure at $u_v$ based on $(\widetilde{\bfN}_*)_{u_v}$.

\noindent\textbf{3.3.4\ \ Assemble}\ \ Each scaffolding tree corresponds to multiple molecules due to the multiple ways substructures can be combined. 
We enumerate all the possible molecules following~\cite{jin2018junction} (See Section~\ref{sec:assemble} in Appendix for more details) for the further selection as described below.

\subsection{Molecule Diversification}
\label{sec:dpp}

In the current iteration, we have generated $M$ molecules ($X_1, \cdots, X_{M}$) and need to select $C$ molecules for the next iteration. 
We expect these molecules to have desirable chemical properties (high $F$ score) and simultaneously maintain higher structural diversity. 
To do so, we resort to the \textit{determinantal point process (DPP)}~\cite{kulesza2012determinantal}. 
DPP models the repulsive correlation between data points~\cite{kulesza2012determinantal} and has been successfully applied to many applications such as text summarization~\cite{Cho19dpp}, mini-batch sampling~\cite{zhang2017determinantal}, and recommendation system~\cite{chen2018fast}. 
For $M$ data points, whose indexes are $\{1,2,\cdots, M\}$, $\bfS \in \RB_{+}^{M\times M}$ denotes the similarity kernel matrix between these data points. To create a diverse subset (denoted $\calR$) with fixed size $C$, the sampling probability should be proportional to the determinant of the submatrix $\bfS_{\calR} \in \RB^{C\times C}$, i.e.,  
$P(\calR) \propto \det(\bm{S}_{\calR})$, {where} $\calR\subseteq \{1,2,\cdots, M\},\ |\calR|=C$.  
Combining the objective ($F$) value and diversity, the composite objective is  
\begin{equation}
\label{eqn:dpp_obj}
\underset{\calR \subseteq\{1,2,\cdots, M\}, |\calR|=C}{\arg\max}\ \calL_{\text{DPP}}(\calR) = \lambda \sum_{r\in \calR} F(X_r) + \log P({\calR}) = \log \det(\bm{V}_{\calR}) + \log \det(\bm{S}_{\calR}), 
\end{equation}
where the hyperparamter $\lambda > 0$ balances the two terms, the diagonal scoring matrix:
\begin{equation}
\bm{V} = \text{diag}\Big( \big[ \exp(\lambda F(X^{}_1)), \cdots, \exp(\lambda F(X^{}_{M})) \big] \Big)
\end{equation}
$\bm{V}_{\calR}\in \RB^{C\times C}$ is a sub-matrix of $\bfV$ indexed by $\calR$. When $\lambda$ goes to infinity, it is equivalent to selecting $C$ candidates with the highest $F$ score regardless of diversity, same as conventional evolutionary learning in~\cite{jensen2019graph,nigam2019augmenting}. 
Inspired by generalized DPP methods~\cite{kulesza2012determinantal,chen2018fast}, we further transform $\calL_{\text{DPP}}(\calR)$, 
\begin{equation}
\label{eqn:dpp_obj2}
\begin{aligned}
\calL_{\text{DPP}}(\calR) & = \log \det(\bm{V}_{\calR}) + \log \det(\bm{S}_{\calR}) = \log \det(\bm{V}_{\calR} \bm{S}_{\calR})  \\
& = \log \det \Big( \bm{V}_{\calR}^{\frac{1}{2}} \bm{S}_{\calR} \bm{V}_{\calR}^{\frac{1}{2}} \Big) 
= \log \det \Big( \big( \bm{V}^{\frac{1}{2}} \bfS \bm{V}^{\frac{1}{2}} \big)_{\calR} \Big).   
\end{aligned}
\end{equation}
where $\bm{V}^{\frac{1}{2}} \bfS \bm{V}^{\frac{1}{2}}$ is symmetric positive semi-definite. Then it can be solved by generalized DPP methods in $O(C^2M)$~\cite{chen2018fast} (Section~\ref{sec:diversity_theory} in Appendix). 
The computational complexity of \mname is $O(TMC^2)$ (see Section~\ref{sec:complexity} in Appendix). 
Algorithm~\ref{alg:main} summarizes the entire algorithm.

\begin{algorithm}[h!]
\caption{Differentiable Scaffolding Tree (\mname)} 
\label{alg:main}
\begin{algorithmic}[1]
\STATE \textbf{Input}: Iteration $T$, population size $C$, input molecule $X^{(1)}_{}$. Initial population $\Phi = \{X^{(1)}_{}\}$.
\STATE
\textbf{Output}: Generated Molecule Set $\Omega$. 
\STATE Learn GNN (Eq.~\ref{eqn:gnn_objective}): $\Theta_{*} = {\arg\min}_{\Theta} \sum_{(X,y)\in \mathcal{D}}^{} \calL(y, \widehat{y})$. \ \ \ \# Section~\ref{sec:gnn}. 
\FOR{$t=1,2,\cdots, T$}
\STATE Initialize set $\Gamma = \{\}$. 
\FOR{$X^{(t)} \in \Phi$}
\STATE Initialize \DST $\{ \widetilde{\bfN}, \widetilde{\bfA}, \widetilde{\bfw}\}$ for $X^{(t)}$ (Eq.~\ref{eqn:differentiable_node_index_matrix},~\ref{eqn:diff_adj},~\ref{eqn:diff_node_weight}). 
\STATE Optimize \DST: $\bfN_*, \bfA_*, \bfw_* = {\arg\max}_{ \{ \widetilde{\bfN}, \widetilde{\bfA}, \widetilde{\bfw} \} } \ \ \text{GNN}(\{\widetilde{\bfN}, \widetilde{\bfA}, \widetilde{\bfw} \}; \Theta_{*})$ (Eq.\ref{eqn:optimize}). 
\STATE Sample from \DST: $\calT^{(t+1)}_{j} \sim \text{DST-Sampler}\big(\bfN_*, \bfA_*, \bfw_* \big), j=1,2,\cdots$ (Eq.\ref{eqn:sample}); Assemble scaffolding tree $\calT^{(t+1)}_{j}$ into molecules $X_{j}^{(t+1)}$. 
\STATE $\Gamma = \Gamma\cup \{X^{(t+1)}_j\}$. 
\ENDFOR 
\STATE Select $\Phi \subseteq \Gamma, |\Phi|=C$ based on Eq.~\ref{eqn:dpp_obj}; \ \  $\Omega = \Omega \cup \Phi$. \ \ \ \# Section~\ref{sec:dpp}
\ENDFOR
\end{algorithmic}
\end{algorithm}

\section{Experiment}
\label{sec:experiment}


\subsection{Experimental Setup}
\label{sec:setup}

\noindent\textbf{Molecular Properties} contains \textbf{QED}; 
\textbf{LogP}; 
\textbf{SA}; 
\textbf{JNK3}; 
\textbf{GSK3\bm{$\beta$}}, \marked{following~\cite{jin2020multi,nigam2019augmenting,moss2020boss,xie2021mars}, where QED quantifies drug-likeness; LogP indicates the water-octanol partition coefficient; SA stands for synthetic accessibility and is used to prevents the formation of chemically unfeasible molecules; JNK3/GSK3$\bm{\beta}$ measure inhibition against c-Jun N-terminal kinase-3/Glycogen synthase kinase 3 beta. }
For all the 5 scores (including normalized SA), higher is better. 
We conducted (1) single-objective generation that optimizes JNK3, GSK3\bm{$\beta$} and LogP separately
and (2) multi-objective generation that optimizes the mean value of  ``JNK3+GSK3\bm{$\beta$}'' and ``QED+SA+JNK3+GSK3\bm{$\beta$}'' in the main text. 
Details are in Section~\ref{sec:property}.

\noindent\textbf{Dataset}: ZINC 250K contains around 250K druglike molecules~\cite{sterling2015zinc}. 
We select the substructures that appear more than 1000 times in ZINC 250K as the vocabulary set $\calS$, which contains 82 most frequent substructures. Details are in Section~\ref{sec:dataset}. 

\noindent\textbf{Baselines}. 
(1) \textbf{LigGPT} (string-based distribution learning model with Transformer as a decoder)~\cite{bagal2021liggpt}; 
(2) \textbf{GCPN} (Graph Convolutional Policy Network)~\cite{You2018-xh}; 
(3) \textbf{MolDQN} (Molecule Deep Q-Network)~\cite{zhou2019optimization}; 
(4) \textbf{GA+D} (Genetic Algorithm with Discriminator network)~\cite{nigam2019augmenting};  
(5) \textbf{MARS} (Markov Molecular Sampling)~\cite{xie2021mars}; 
(6) \textbf{RationaleRL}~\cite{jin2020multi}; 
\marked{
(7) \textbf{ChemBO} (Chemical Bayesian Optimization)~\cite{korovina2020chembo}; 
(8) \textbf{BOSS} (Bayesian Optimization over String Space)~\cite{moss2020boss}. } 
Among them, LigGPT belongs to deep generative model, where all the oracle calls can be precomputed; 
GCPN, MolDQN are deep reinforcement learning methods; 
GA+D, MARS are evolutionary learning methods; 
RationaleRL is deep generative model fine-tuned with RL techniques. 
\marked{ChemBO and BOSS are Bayesian optimization methods. We also consider a \mname variant: \textbf{\mname-rand}. Instead of optimizing and sample from \mname, \mname-rand leverages random local search, i.e., randomly selecting basic operations (EXPAND, REPLACE, SHRINK) and substructure from vocabulary. To improve efficiency, we also select a subset of all the random samples with high surrogate GNN prediction scores. } 
All the baselines except LigGPT require online oracle calls.
Details are in Section~\ref{sec:baseline}. 

\noindent\textbf{Metrics}. 
\marked{For each method, we select top-100 molecules with highest property scores for evaluation, and consider the following metrics following~\cite{jin2018junction,You2018-xh,jin2020multi,xie2021mars}}
(1) \textbf{Novelty (Nov)} \marked{(\% of the generated molecules that are not in training set)}; 
(2) \textbf{Diversity (Div)} \marked{(average pairwise Tanimoto distance between the Morgan fingerprints)}; 
(3) \marked{\textbf{Average Property Score (APS)} (average score of top-100 molecules)};
(4) \textbf{\# of oracle calls}: \mname needs to call oracle in labeling data for GNN (\textbf{precomputed}) and \mname based \textit{de novo} generation (\textbf{online}), we show the costs for both steps. \marked{For each method in Table~\ref{table:denovo_multi} and~\ref{table:denovo_single}, we set the number of oracle calls so that the property score nearly converge w.r.t. oracle call’s number. }
Details are in Section~\ref{sec:metrics}.

\begin{table*}[h!]
\tiny
\parbox{.25\textwidth}{\caption{Multi-objective \textit{de novo} design. \#oracle = (1)``\textbf{precomputed} oracle call'' (to label molecules in existing database) + (2)``\textbf{online} oracle call'' (during learning). }
\label{table:denovo_multi}
}
\hspace{0.03\textwidth}
\parbox{.60\textwidth}{
\centering
\begin{tabular}{lcccc|cccc}
\toprule[0.6pt]
\multirow{2}{*}{Method} & \multicolumn{4}{c}{JNK3+GSK3\bm{$\beta$}} & \multicolumn{4}{c}{QED+SA+JNK3+GSK3\bm{$\beta$}} \\ 
& Nov$\uparrow$ & Div$\uparrow$ & {APS}$\uparrow$ & \#oracle$\downarrow$ & Nov$\uparrow$ & Div$\uparrow$ & {APS}$\uparrow$ & \#oracle$\downarrow$ \\
\hline 
LigGPT & \textbf{100\%} & \textbf{0.845} & 0.271 & 100k+0 & \textbf{100\%} & \textbf{0.902} & 0.378 & 100k+0 \\ 
GCPN & \textbf{100\%} & 0.578 & 0.293 & 0+200K & \textbf{100\%} & 0.596 & 0.450 & 0+200K \\ 
MolDQN & \textbf{100\%} & 0.605 & 0.348 & 0+200K & \textbf{100\%} & 0.597 & 0.365 & 0+200K \\ 
GA+D & \textbf{100\%} & 0.657 & 0.608 & 0+50K & 97\% & 0.681 & 0.632 & 0+50K \\ 
RationaleRL & \textbf{100\%} & 0.700 & 0.795 & 25K+67K & 99\% & 0.720 & 0.675 & 25K+67K \\ 
MARS & \textbf{100\%} & 0.711 & 0.789 & 0+50K  & \textbf{100\%} & 0.714 & 0.662 & 0+50K \\ 
{ChemBO} & 98\% & 0.702 & 0.747 & 0+50K & 99\% & 0.701 & 0.648 & 0+50K \\ 
{BOSS} & 99\% & 0.564 & 0.504 & 0+50K & 98\% & 0.561 & 0.504 & 0+50K \\ 
{\mname-rand} & \textbf{100\%} & 0.456 & 0.622 & 10+5K & \textbf{100\%} & 0.765 & 0.575 & 20K+5K \\ 
\mname & \textbf{100\%} &  0.750 & \textbf{0.827} & \textbf{10K+5K} & \textbf{100}\% &  0.755 & \textbf{0.752} & \textbf{20K+5K} \\ 
\bottomrule[0.6pt]
\end{tabular}}
\end{table*}


\subsection{Optimization Performance}
\label{sec:denovo}

The results of multi-objective and single-objective generation are shown in Table~\ref{table:denovo_multi} and~\ref{table:denovo_single}. 
We find that DGM (LigGPT) and RL based methods (GCPN and MolDQN) fails in some tasks, which is consistent with the results reported in RationaleRL~\cite{jin2020multi} and MARS~\cite{xie2021mars}. 
Overall, \mname obtains the best results in most tasks. 
In terms of success rate and diversity, \mname outperformed all baselines in most tasks. 
It also reached the highest scores within $T=50$ iterations in most optimization tasks (see Table \ref{table:denovo_top} and \ref{table:denovo_multi_top} in Appendix). 
Especially in optimizing LogP, the model successfully learned to add a six-member ring (see Figure~\ref{fig:logp_growing} in Appendix) each step, which is theoretically the optimal strategy under our setting. 
Combined with the ablation study comparing with random selection (see Figure \ref{fig:ablation} in Appendix), our results show the local gradient defined by \mname is a useful direction indicator, consistent with the concept of gradient. 
Further, achieving high diversity validates the effect of the DPP-based selection strategy. 
Although the novelty is not the highest, it is still comparable to baseline methods. These results show our gradient-based optimization strategy has a strong optimization ability to provide a diverse set of molecules with high objective functions.

\begin{table*}[h!]
\tiny
\centering
\caption{Single-objective \textit{de novo} molecular generation.  }
\label{table:denovo_single}
 \resizebox{1.0\textwidth}{!}{
\begin{tabular}{lcccc|cccc|cccc}
\toprule[0.6pt]
\multirow{2}{*}{Method} & \multicolumn{4}{c}{JNK3} & \multicolumn{4}{c}{GSK3\bm{$\beta$}} & \multicolumn{4}{c}{LogP} \\ 
& Nov$\uparrow$ & Div$\uparrow$ & {APS}$\uparrow$ & \#oracle$\downarrow$ & Nov$\uparrow$  & Div$\uparrow$ & {APS}$\uparrow$ & \#oracle$\downarrow$ & Nov$\uparrow$ & Div$\uparrow$ & {APS}$\uparrow$ & \#oracle$\downarrow$ \\
\hline 
LigGPT & \textbf{100\%} & \textbf{0.837} & 0.302 & 100K+0 & \textbf{100\%} & \textbf{0.867} & 0.283 & 100K+0 & \textbf{100\%} & \textbf{0.868} & 4.56 & 100K+0 \\ 
GCPN & \textbf{100\%} & 0.584 & 0.365 & 0+200K  & \textbf{100\%} & 0.519 & 0.400 & 0+200K  & \textbf{100\%} & 0.532 & 5.43 & 0+200K \\
MolDQN & \textbf{100\%} & 0.605 & 0.459  & 0+200K & \textbf{100\%} & 0.545 & 0.398 & 0+200K  & \textbf{100\%} & 0.485 & 6.00 & 0+200K  \\
GA+D & 99\% & 0.702 & 0.615  & 0+50K & 98\% & 0.687 & 0.678 & 0+50K & \textbf{100\%} & 0.721 & 30.2 & 0+50K \\
RationaleRL & 99\% & 0.681 & 0.803 & 25K+32K & 99\% & 0.731 & 0.806 & 30K+45K & - & - & -  & - \\
MARS & \textbf{100\%} & 0.711 & 0.784  & 0+50K & \textbf{100\%} & 0.735 & 0.810 & 0+50K & \textbf{100\%} & 0.692 & 44.1 & 0+30K \\ 
{ChemBO} & 98\% & 0.645 & 0.648 & 0+50K & 98\% & 0.679 & 0.492 & 0+50K & 98\% & 0.732 & 10.2 & 0+50K \\ 
{BOSS} & 98\% & 0.601 & 0.471 & 0+50K & 99\% & 0.658 & 0.432 & 0+50K & 100\% & 0.735 & 9.64 & 0+50K  \\ 
{\mname-rand} & \textbf{100\%} & 0.754 & 0.413 & 10K+10K & 97\% & 0.793 & 0.455 & 10K+10K & \textbf{100\%} & 0.713 & 36.1 & 10K+15K \\ 
\mname & \textbf{100\%} & 0.732 & \textbf{0.928}  & \textbf{10K+5K} & \textbf{100\%} & 0.748 & \textbf{0.869}  & \textbf{10K+5K} & \textbf{100\%} & 0.704 & \textbf{47.1} & \textbf{10K+5K} \\  
\bottomrule[0.6pt]
\end{tabular}}
\end{table*}

\subsection{Oracle Efficiency}
\label{sec:oracle_experiment}
\begin{figure}[h!]
\centering
{\includegraphics[width=0.86\linewidth]{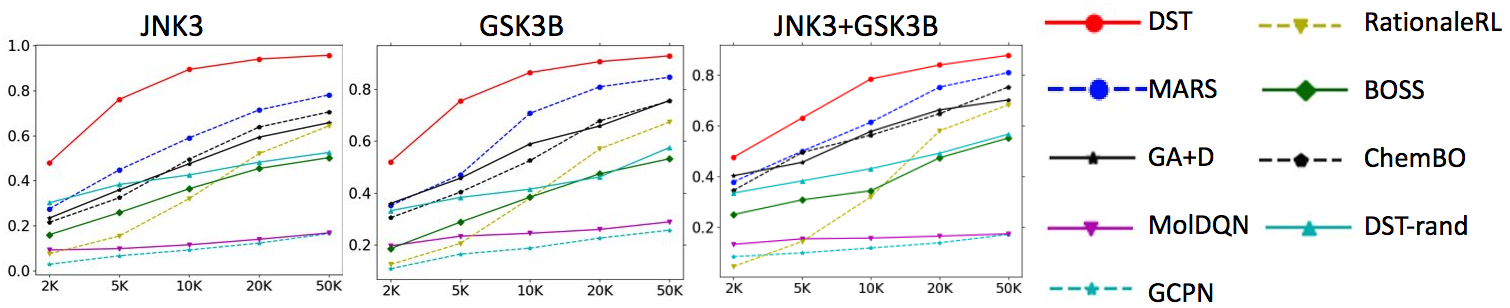}}
\caption{\marked{
Oracle efficiency test. Top-100 average score v.s. number of oracle calls. }
}
\label{fig:oracle_num}
\end{figure}

As mentioned above, oracle calls for realistic optimization tasks can be time-consuming and expensive. 
From Table~\ref{table:denovo_multi} and~\ref{table:denovo_single}, we can see that majority of de novo optimization methods require oracle calls online (instead of precomputation), including all of RL/evolutionary algorithm based baselines. \mname takes fewer oracle calls compared with baselines. \mname can leverage the precomputed oracle calls to label the molecules in an existing database (i.e., ZINC) for training the oracle GNN and dramatically saving the oracle calls during reference. 
In the three tasks in Table~\ref{table:denovo_single}, two-thirds of the oracle calls (10K) can be precomputed or collected from other sources. 
To further verify the oracle efficiency, we explore a special setting of molecule optimization where the budget of oracle calls is limited to a fixed number (2K, 5K, 10K, 20K, 50K) and compare the optimization performance.
For GCPN, MolDQN, GA+D and MARS, the learning iteration number depends on the budget of oracle calls. 
RationaleRL~\cite{jin2020multi} is not included because it requires intensive oracle calls to collect enough reference data, exceeding the oracle budget in this scenario. 
In \mname, we use around 80\% budget to label the dataset (i.e., training GNN) while the remaining budget to conduct \textit{de novo} design. Specifically, for 2K, 5K, 10K, 20K, 50K, we use 1.5K, 4K, 8K, 16K and 40K oracle calls to label the data for learning GNN, respectively. 
We show the average objective values of top-100 molecules under different oracle budgets in Figure~\ref{fig:oracle_num}. 
Our method shows a significant advantage compared to all the baseline methods in all limited budget settings. 
We conclude the reason as supervised learning is a well-studied and much easier task than generative modeling. 

\subsection{Interpretability Analysis}
\label{sec:interpret}

\begin{figure}[h!]
\centering
\subfigure{
\includegraphics[width=0.6\linewidth]{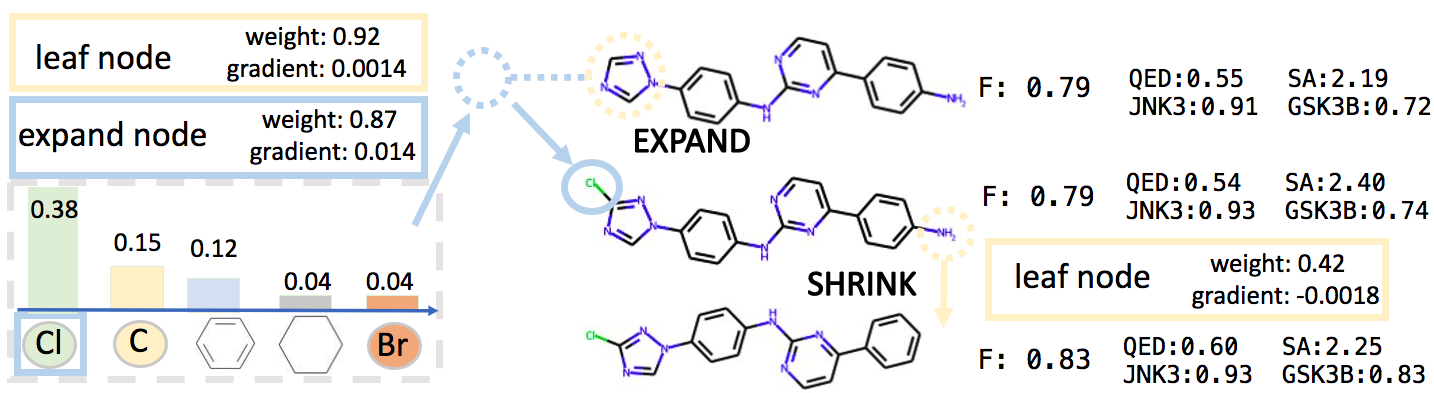}}
\subfigure{
\includegraphics[width=0.3\linewidth]{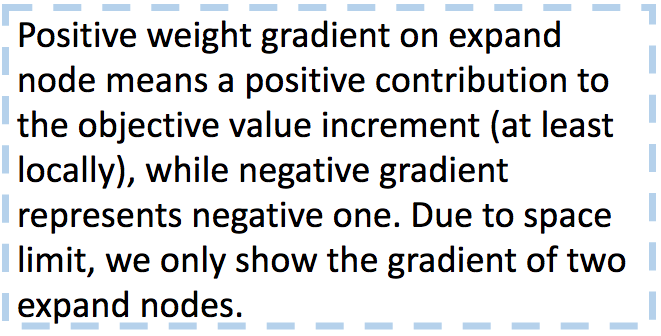}}
\caption{Two steps in optimizing ``QED+SA+JNK3+GSK3\bm{$\beta$}''.  
}
\label{fig:interpret}
\end{figure}


To obtain more insights from the local gradient, We visualize two modification steps in Figure~\ref{fig:interpret}. 
The node weights and their gradient values interpret the property at the substructure level. 
This is similar to most attribution-based interpretable ML methods, e.g., saliency map~\cite{simonyan2013deep}.




\section{Conclusion}
\label{sec:conclu}

This paper proposed \fullname (\mname) to make a molecular graph locally differentiable, allowing a continuous gradient-based optimization. To the best of our knowledge, it is the first attempt to make the molecular optimization problem differentiable at the substructure level, rather than resorting to latent spaces or using RL/evolutionary algorithms. 
We constructed a general molecular optimization strategy based on \mname, corroborated by thorough empirical studies.

\bibliographystyle{plain}
\bibliography{ref}

\newpage 
\appendix

\tableofcontents

\section*{Appendix to Differentiable Scaffolding Tree for Molecular Optimization}

The appendix is organized as follows. 
First, we list the complete mathematical notations for ease of exposition. 
Then, we show additional experimental setup and empirical results, including baseline setup in Section~\ref{sec:baseline}, implementation details of our method in Section~\ref{sec:implementation_details}, additional experimental results in Section~\ref{sec:additional_experiment}. 
Then, we provide theoretical analysis in Section~\ref{sec:additional_theory}, extend molecule diversification in Section and proof of theoretical results in the main paper in Section~\ref{sec:proof}.

\begin{table}[h!]
\small 
\centering
\caption{Complete Mathematical Notations.}
\resizebox{1.0\columnwidth}{!}{
\begin{tabular}{c|l}
\toprule[1pt]
Notations & Descriptions \\ 
\hline 
$\calO$ & Oracle function, e.g., evaluator of molecular property (Def~\ref{def:oracle}).  \\
$F$ & objective function of molecule generation (Eq.~\ref{eqn:objective}). \\ 
$P\in \mathbb{N}_{+}$ & Number of target oracles. \\  
$\calQ$ & Set of all the valid chemical molecules. \\ 
$\calS$ & Vocabulary set, i.e., substructure set. A substructure is an atom or a ring. \\ 
$\calT$ & Scaffolding tree (Def~\ref{def:tree}). \\ 
$K = |\calT| $ & number of nodes in scaffolding tree $\calT$.   \\ 
$\bfN; \bfA; \bfw$ & Node indicator matrix; adjacency matrix; node weight. \\ 
$\calV_{\text{leaf}}$ & Leaf node set in scaffolding tree $\calT$. \\
$\calV_{\text{nonleaf}}$ & Nonleaf node set in scaffolding tree $\calT$. \\ 
$\calV_{\text{expand}}$ & Expansion node set in scaffolding tree $\calT$. \\ 
$K_{\text{leaf}} = |\calV_{\text{leaf}}|$ & Size of leaf node set.  \\  
$K_{\text{expand}} = |\calV_{\text{expand}}| = K$ & Size of expansion node set. $K_{\text{leaf}} = K_{\text{expand}}$.  \\  \hline 
$d\in \mathbb{N}_{+} $ & GNN hidden dimension. \\ 
$L \in \mathbb{N}_{+}$ & GNN depth. \\
$\Theta = \{\bfE\}\cup \{\bfB^{(l)}, \bfU^{(l)}\}_{l=1}^{L}$ & Learnable parameter of GNN. \\ 
$\bfE \in \RB^{|\calS|\times d}$ & embedding stackings of all the substructures in vocabulary set $\calS$. \\ 
$\bfB^{(l)} \in \RB^{K\times d}$ & bias parameters at $l$-th layer. \\
$\bfU^{(l)} \in \RB^{d\times d}$ & weight parameters at $l$-th layer. \\
$\bfH^{(l)}, l=0,\cdots,L$ & Node embedding at $l$-th layer of GNN\\ 
$\bfH^{(0)} = \bfN \bfE \in \RB^{K\times d}$ & initial node embeddings, stacks basic embeddings of all the nodes in the scaffolding tree. \\ 
MLP & multilayer perceptron \\
ReLU & ReLU activate function \\ 
$\haty$ & GNN prediction. \\ 
$y$ & groundtruth\\
$\calL$ & Loss function of GNN. \\
$\mathcal{D}$ & the training set \\
$\calN(X)$ & Neighborhood molecule set of $X$ (Def~\ref{def:neighbor}). \\ 
$\Lambda$ & differentiable edge set. \\
$\widetilde{\bfN}; \widetilde{\bfA}; \widetilde{\bfw}$ & Differentiable node indicator matrix; adjacency matrix; node weight. \\ 
 \hline 
$\det()$ & Determinant of a square matrix \\ 
$M \in \mathbb{N}_{+}$ & Number of all possible molecules to select. \\
$C \in \mathbb{N}_{+}$ & Number of selected molecules.  \\ 
$\bfS \in \RB^{M\times M}_{+}$ & Similarity kernel matrix.  \\ 
$\bfV \in \RB^{M\times M}_{+}$ & Diagonal scoring matrix. \\ 
$\calR$ & subset of $\{1,2,\cdots, M\}$, index of select molecules. \\ 
$\lambda > 0$ &  hyperparameter in Eq.~\ref{eqn:dpp_obj} and ~\ref{eqn:dpp_obj_appendix}, balances desirable property and diversity. \\  
\bottomrule[1pt]
\end{tabular}
}
\label{table:notation_complete}
\end{table}

\section{Complete Mathematical Notations.}
\label{sec:notation}

In this section, we show all the mathematical notations in Table~\ref{table:notation_complete} for completeness.

\section{Baseline Setup}
\label{sec:baseline}

In this section, we describe the experimental setting for baseline methods. 
Most of the settings follow the original papers. 
\begin{itemize}
\item \textbf{LigGPT} (string-based distribution learning model with Transformer as a decoder)~\cite{bagal2021liggpt} is trained for 10 epochs using the Adam optimizer with a learning rate of $6e-4$. 
LigGPT comprises stacked decoder blocks, each of which, is composed of a masked self-attention layer and fully connected neural network. 
Each self-attention layer returns a vector of size 256, that is taken as input by the fully connected network. The hidden layer of the neural network outputs a vector of size 1024 and uses a GELU activation and the final layer again returns a vector of size 256 to be 7 used as input for the next decoder block. LigGPT consists of 8 such decoder blocks. LigGPT has around 6M parameters. 
\item \textbf{GCPN} (Graph Convolutional Policy Network)~\cite{You2018-xh} leveraged graph convolutional network and policy gradient to optimize the reward function that incorporates target molecular properties and adversarial loss. 
In each step, the allowable action to the current molecule could be either connecting a new substructure or an atom with an existing molecular graph or adding a bond to connect existing
atoms. GCPN predicts the actions and is trained via proximal policy optimization (PPO) to optimize an accumulative reward, including molecular property objectives and adversarial loss. 
Both policy network and adversarial network (discriminative training) use the same neural architecture, which is a three-layer graph convolutional network (GCN)~\cite{kipf2016semi} with 64 hidden nodes. Batch normalization is adopted after each layer, and sum-pooling is used as the aggregation function. Adam optimizer is used with 1e-3 initial learning rate, and batch size is 32.

\item \textbf{MolDQN} (Molecule Deep Q-Networks)~\cite{zhou2019optimization}, same as GCPN, formulate the molecule generation procedure as a Markov Decision Process (MDP) and use Deep Q-Network to solve it. 
The reward includes target property and similarity constraint. 
Following the original paper, the episode number is 5,000, maximal step in each episode is 40. Each step calls oracle once; thus, 200K oracle calls are needed in one generation process. 
The discount factor is 0.9. 
Deep Q-network is a multilayer perceptron (MLP) whose hidden dimensions are 1024, 512, 128, 32, respectively. The input of the Q-network is the concatenation of the molecule feature (2048-bit Morgan fingerprint, with a radius of 3) and the number of left steps. 
Adam is used as an optimizer with 1e-4 as the initial learning rate.
Only rings with a size of 5 and 6 are allowed. 
It leverages $\epsilon$-greedy together with randomized value functions (bootstrapped-DQN) as an exploration policy, $\epsilon$ is annealed from 1 to 0.01 in a piecewise linear way. 
\item \textbf{GA+D} (Genetic Algorithm with Discriminator network)~\cite{nigam2019augmenting} uses a deep neural network as a discriminator to enhance exploration in a genetic algorithm. 
$\beta$ is an important hyperparameter that weights the importance of the discriminator's loss in the overall fitness function, and we set it to 10. 
The generator runs 100 generations with a population size of 100 for \textit{de novo} molecular optimization and 50 generations with a population size of 50 for molecular modification. 
Following the original paper~\cite{nigam2019augmenting}, the architecture of the discriminator is a two-layer fully connected neural network with ReLU activation and a sigmoid output layer. The hidden size is 100, while the size of the output layer is 1. 
The input feature is a vector of chemical and geometrical properties characterizing the molecules. 
We used Adam optimizer with 1e-3 as the initial learning rate. 
\item \textbf{RationaleRL}~\cite{jin2020multi} is a deep generative model that grows a molecule atom-by-atom from an initial rationale (subgraph). 
The architecture of the generator is a message-passing network (MPN) followed by MLPs applied in breadth-first order. The generator is pre-trained on general molecules combined with an encoder and then fine-tuned to maximize the reward function using policy gradient. 
The encoder and decoder MPNs both have hidden dimensions of 400. The dimension of the latent variable is 20. Adam optimizer is used on both pre-training and fine-tuning with initial learning rates of 1e-3, 5e-4, respectively. The annealing rate is 0.9. We pre-trained the model with 20 epochs. 


\item \textbf{MARS}~\cite{xie2021mars} leverage Markov chain Monte Carlo sampling (MCMC) on molecules with an annealing scheme and an adaptive proposal. The proposal is parameterized by a graph neural network, which is trained on MCMC samples. 
We follow most of the settings in the original paper. 
The message passing network has six layers, where the node embedding size is set to 64. 
Adam is used as an optimizer with 3e-4 initial learning rate. To generate a basic unit, top-1000 frequent fragments are drawn from ZINC database~\cite{sterling2015zinc} by enumerating single bonds to break.  
During the annealing process, the temperature $T = 0.95^{\lfloor t/5\rfloor}$ would gradually decrease to 0. 
\end{itemize}

\section{Implementation Details}
\label{sec:implementation_details}

\subsection{Dataset}
\label{sec:dataset}
We use ZINC 250K dataset, which contains around 250K druglike molecules extracted from the ZINC
database~\cite{sterling2015zinc}. 
The clean data is available at \cite{huang2021therapeutics,du2021graphgt} (\url{https://tdcommons.ai/generation_tasks/molgen/}). 
We first clean the data by removing the molecules containing out-of-vocabulary substructure and having 195K molecules left.

\begin{figure}[h!]
\centering
\includegraphics[width=12.8cm]{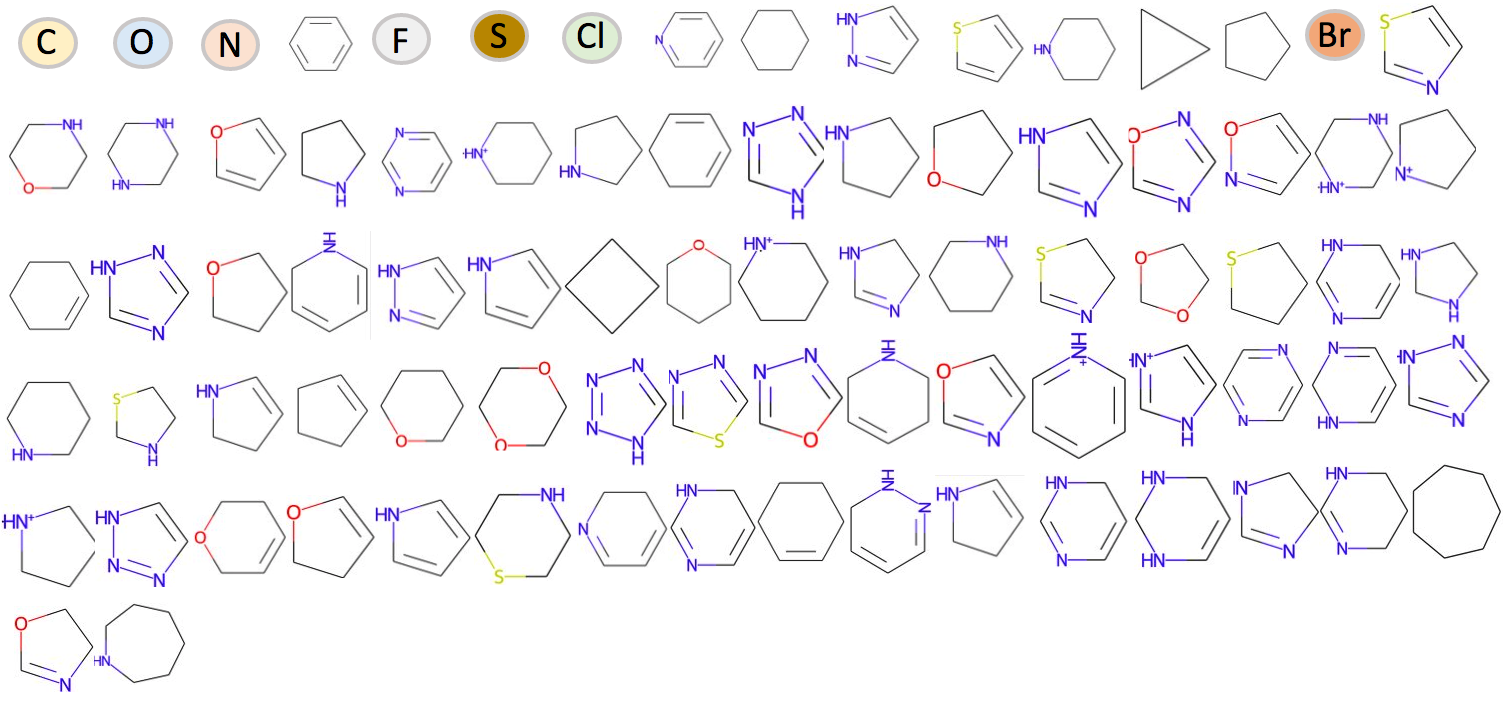}
\caption{All the substructures in the vocabulary set $\calS$, drawn from ZINC 250K database~\cite{sterling2015zinc}. It includes atoms and single rings appearing more than 1000 times in the ZINC250K database. 
}
\label{fig:vocab}
\end{figure}

\textbf{Vocabulary $\calS$: set of substructure}. 
The substructure is the basic building block in our method, including frequent atoms and rings. 
On the other hand, atom-wise molecule generation is difficult due to the existence of rings. 
To select the substructure set $\calS$, we break all the ZINC molecules into substructures (including single rings and single atoms), count their frequencies, and include the substructures whose frequencies are higher than 1000 into vocabulary set $\calS$. 
The final vocabulary contains 82 substructures, including the frequent atoms like carbon atom, oxygen atom, nitrogen atom, and frequent rings like benzene ring. 
The vocabulary size is big enough for this proof-of-concept study. 
Other works also need to constrain their design space, such as MolDQN only allowing three types of atoms in a generation: ``C'', ``N'', ``O''~\cite{zhou2019optimization};  JTVAE~\cite{jin2018junction,jin2019learning}, as well as RationaleRL~\cite{jin2020multi} only using frequent substructures similar to our setting. 
On the other hand, we may not want infrequent atoms or substructures because rare substructures in ZINC may have some undesired properties such as toxicity, may not be stable, may not be easily synthesizable~\cite{gao2020synthesizability}. 
Also, rare substructures may impede the learning of oracle GNN. 
Note that users can enlarge the substructure space when they apply our method. 
We show all the 82 substructures in $\calS$ in Figure~\ref{fig:vocab}.


\subsection{Software/Hardware Configuration}
\label{sec:hardware_config}
We implemented \mname using Pytorch 1.7.0, Python 3.7, RDKit v2020.09.1.0 on an Intel Xeon E5-2690 machine with 256G RAM and 8 NVIDIA Pascal Titan X GPUs.

\subsection{Target molecular properties}
\label{sec:property}
{Target molecular properties} include
\begin{itemize}
\item \textbf{QED} represents a quantitative estimate of drug-likeness. QED score ranges from 0 to 1. It can be evaluated by the RDKit package (\url{https://www.rdkit.org/}). 
\item \textbf{LogP} represents octanol-water partition coefficient, measuring molecules' solubility. LogP score ranges from $-\infty$ to $+\infty$. Thus, when optimizing LogP individually, we use the GNN model to do regression. 
\item \textbf{SA} (Synthetic Accessibility) score measures how hard it is to synthesize a given molecule, based on a combination of the molecule’s fragments contributions~\cite{ertl2009estimation}. 
It is evaluated via RDKit~\cite{landrum2006rdkit}. 
The raw SA score ranges from 1 to 10. A higher SA score means the molecule is hard to be synthesized and is not desirable. 
In the multiple-objective optimization, we normalize the SA score to $[0,1]$ so that a higher normalized SA value mean easy to synthesize. 
Following~\cite{gao2020synthesizability}, we use the normalize function for raw SA score, 
\begin{equation*}
\text{normalized-SA}(X)=
\left\{
\begin{array}{ll}
1,  & \text{SA}(X) < \mu \\
\exp{(-\frac{(\text{SA}(X) - \mu)^2}{2\sigma^2})},  &  \text{SA}(X) \geq \mu, \\
\end{array}
\right. 
\end{equation*}
where $\mu = 2.230044, \sigma = 0.6526308$. 
\item \textbf{JNK3} (c-Jun N-terminal Kinases-3) belongs to the mitogen-activated protein kinase family and are responsive to stress stimuli, such as cytokines, ultraviolet irradiation, heat shock, and osmotic shock. Similar to GSK3\bm{$\beta$}, JNK3 is also evaluated by well-trained\footnote{The test AUROC score is 0.86~\cite{jin2020multi}. } random forest classifiers using ECFP6 fingerprints using ExCAPE-DB dataset~\cite{li2018multi,jin2020multi}, and the range is also $[0,1]$. 
\item \textbf{GSK3\bm{$\beta$}} (Glycogen synthase kinase 3 beta) is an enzyme that in humans is encoded by the GSK3\bm{$\beta$} gene. Abnormal regulation and expression of GSK3\bm{$\beta$} is associated with an increased susceptibility towards bipolar disorder. It is evaluated by well-trained\footnote{The test AUROC score is also 0.86~\cite{jin2020multi}. }  random forest classifiers using ECFP6 fingerprints using ExCAPE-DB dataset~\cite{li2018multi,jin2020multi}. GSK3\bm{$\beta$} score of a molecule ranges from 0 to 1. 
\end{itemize}
For QED, LogP, normalized SA, JNK3, and {GSK3\bm{$\beta$}}, higher scores are more desirable under our experimental setting.

\subsection{Evaluation metrics}
\label{sec:metrics}
We leverage the following {evaluation metrics} to measure the optimization performance:  
\begin{itemize}
\item \textbf{Novelty} is the fraction of the generated molecules that do not appear in the training set. 
\item  \textbf{Diversity} of generated molecules is defined as the average pairwise Tanimoto distance between the Morgan fingerprints~\cite{You2018-xh,jin2020multi,xie2021mars}. 
\begin{equation}
\text{diversity} = 1 - \frac{1}{|\calZ|(|\calZ|-1)}\sum_{Z_1,Z_2 \in \calZ} \text{sim}(Z_1,Z_2),
\end{equation}
where $\calZ$ is the set of generated molecules. $\text{sim}(Z_1,Z_2)$ is the Tanimoto similarity between molecule $Z_1$ and $Z_2$. 
\item \textbf{(Tanimoto) Similarity} measures the similarity between the input molecule and generated molecules. 
It is defined as
\[
\text{sim}(X,Y) = \frac{\bfb_{X}^\top \bfb_{Y}}{\Vert \bfb_{X}\Vert_2 \Vert \bfb_{Y}\Vert_2}, 
\] 
$\bfb_{X}$ is the binary Morgan fingerprint vector for the molecule $X$. In this paper, it is a 2048-bit binary vector. 
\item \textbf{SR} (Success Rate) is the percentage of the generated molecules that satisfy the property constraint measured by objective $f$ defined in Equation~\eqref{eqn:objective}.  
For single-objective \textit{de novo} molecular generation, the objective $f$ is the property score, the constraints for JNK3, GSK3\bm{$\beta$} and LogP are JNK3$\geq 0.5$, GSK3\bm{$\beta$}$\geq 0.5$ and LogP$\geq5.0$ respectively. 
For multi-objective \textit{de novo} molecular generation, the objective $f$ is the average of all the normalized target property scores. Concretely, when optimizing ``JNK3+GSK3\bm{$\beta$}'', both JNK3 and GSK3\bm{$\beta$} ranges from 0 to 1, $f$ is average of JNK3 and GSK3\bm{$\beta$} scores; when optimizing ``{QED+SA+JNK3+GSK3\bm{$\beta$}}'', we first normalized SA to 0 to 1. $f$ is average of QED, normalized SA, JNK3 and GSK3\bm{$\beta$} scores. 
The constraint is the $f$ score is greater than 0.4. 
\item \textbf{\# of oracle calls} during the generation process. \mname needs to call oracle in labeling data for GNN and \mname based \textit{de novo} generation, thus we show the costs for both steps. 
\item \textbf{chemical validities}. 
As we only enumerate valid chemical structures during the recovery from scaffolding trees (Section~\ref{sec:assemble}), the chemical validities of the molecules produced by \mname are always 100\%. 
\end{itemize}

\subsection{Assembling Molecule from Scaffolding Tree}
\label{sec:assemble}

\begin{figure}[h!]
\centering
\subfigure[Ring-atom connection. When connecting atom and ring in a molecule, an atom can be connected to any possible atoms in the ring. In the example, there are 4 possible ways to add a Chlorine atom (``Cl'') as an expansion node to the target ring, which is a leaf node in the scaffolding tree. ]{
\includegraphics[width=12cm]{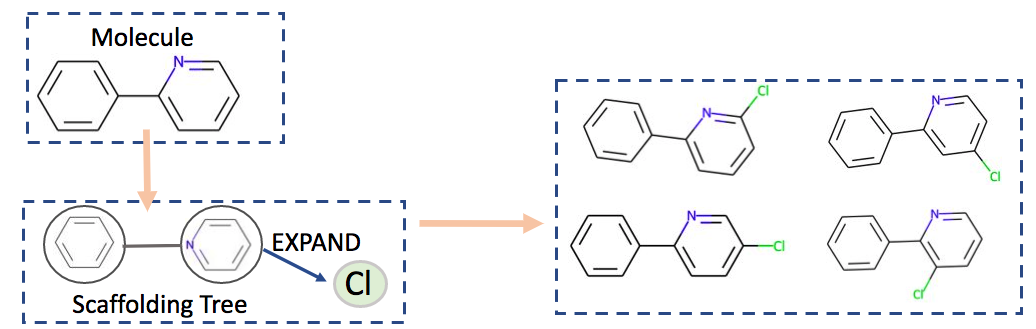}}
\subfigure[Ring-ring connection. When connecting ring and ring, there are two general ways, (1) one is to use a bond (single, double, or triple) to connect the atoms in the two rings. (2) another is two rings share two atoms and one bond. 
In the example, there are 14 possible ways to add a Cyclohexane ring (SMILES is ``C1CCCCC1'') and connect it to the target ring, which is a leaf node in the scaffolding tree. ]{
\includegraphics[width=12.9cm]{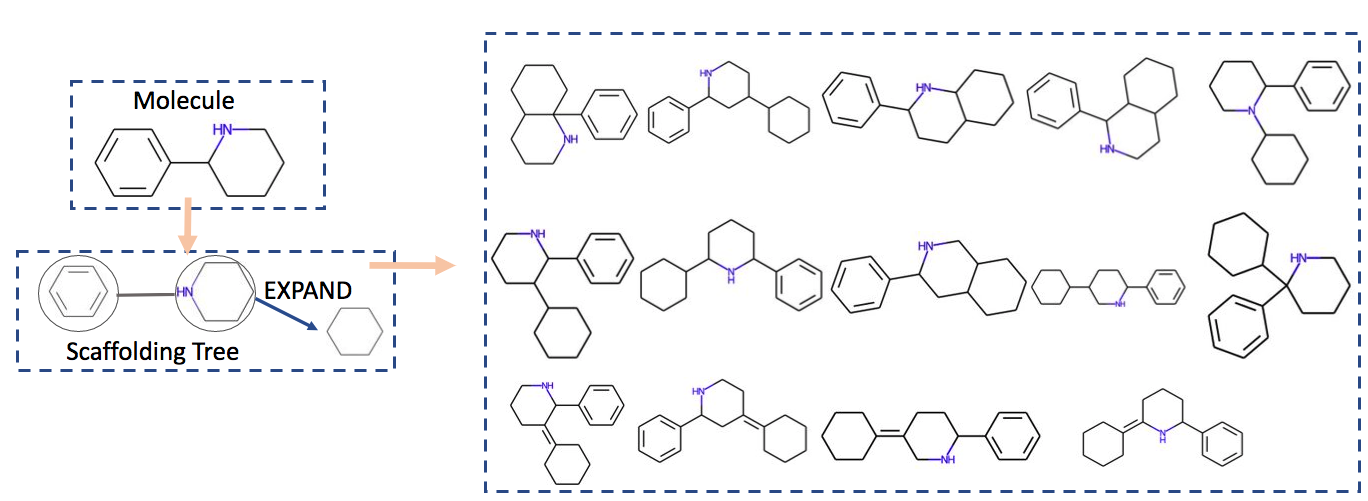}}
\caption{Assemble examples. 
}
\label{fig:assemble}
\end{figure}

Each scaffolding tree corresponds to multiple molecules due to rings' multiple combination ways. For each scaffolding tree, we enumerate all the possible molecules following~\cite{jin2018junction} for further selection. We provide two examples in Figure~\ref{fig:assemble} to illustrate it. Two examples are related to ring-atom combination and ring-ring combination, respectively. For ring-ring combination, our current setting does not support the spiro compounds (contains rings sharing one atom but no bonds) or phenalene-like compounds (contains three rings sharing one atom, and each two of them sharing a bond). These two cases are relatively rare chemical structures in the context of drug discovery~\cite{supsana2005thermal}. 
As we only enumerate valid chemical structures during the recovery from scaffolding trees, the chemical validities are always 100\%.

\subsection{Details on GNN Learning and \mname Optimization}
\label{sec:gnn_setup}

Both the size of substructure embedding and hidden size of GCN (GNN) in Eq.~\eqref{eqn:gcn} are $d=100$. 
The depth of GNN $L$ is 3.  
When training GNN, the training epoch number is 5, and we evaluate the loss function on the validation set every 20K data passes. 
When the validation loss would not decrease, we terminate the training process. 
During the inference procedure, we set the maximal iteration to 5k. 
When optimizing ``JNK3'', ``GSK3\bm{$\beta$}'', ``QED'', ``JNK3+GSK3\bm{$\beta$}'' and ``QED+SA+JNK3+GSK3\bm{$\beta$}'', we use binary cross entropy as loss criterion. 
When optimizing ``LogP'', since LogP ranges from $-\infty$ to $+\infty$, we leverage GNN to conduct regression tasks and use mean square error (MSE) as loss criteria $\calL$. 
In \textit{the de novo} generation, in each generation, we keep $C=10$ molecules for the next iteration. 
In most cases in experiment, the size of the neighborhood set (Definition.~\ref{def:neighbor}) is less than 100. 
We use Adam optimizer with 1e-3 learning rate in training and inference procedure, optimizing the GNN and differentiable scaffolding tree, respectively. 
When optimizing \mname, our method processes one \mname at a time. As a complete generation algorithm, we optimize a batch parallelly and select candidates based on DPP. 
we set the iteration $T$ to a large enough number and tracked the result. When editing cannot improve the objective function or use up oracle budgets, we stop it. All results in the tables are from experiments up to $T=50$ iterations.

\subsection{Results of Different Random Seeds}
\label{sec:error_bar}
In this section, we present the empirical results that use different random seeds for multiple runs. 
In our pipeline, the random error comes from in two steps: 
(1) Training oracle GNN: data selection/split, training process including data shuffle and GNN's parameter initialization. 
(2) Inference (Optimizing \mname): before optimizing \mname, we initialize the learnable parameter randomly, including $\widetilde{\bfN}$, $\widetilde{\bfw}$, $\widetilde{\bfA}$, which also brings randomness.  
To measure the robustness of the proposed method, we use 5 different random seeds for the whole pipeline and compare the difference of 5 independent trials. 
The results are reported in Table~\ref{table:errorbar}. We find that almost all the metrics would not changes significantly among various trials, validating the robustness of the proposed method. 

\begin{table*}[h!]
\tiny
\centering
\caption{Results of 5 independent trials using different random seeds. For novelty, diversity and SR (success rate), we report the average value of 5 runs and their standard deviation.  }
\label{table:errorbar}
 \resizebox{0.6\textwidth}{!}{
\begin{tabular}{ccccc}
\toprule[0.6pt]
Tasks & Novelty$\uparrow$ & Diversity$\uparrow$ & SR$\uparrow$ & \# Oracles$\downarrow$ \\
\hline 
JNK3 & 98.1\%$\pm$0.3\% & 0.722$\pm$0.032 &  92.8\%$\pm$0.5\% & 10K+5K\\ 
GSK3\bm{$\beta$}& 98.6\%$\pm$0.5\% & 0.738$\pm$0.047 & 91.8\%$\pm$0.3\% & 10K+5K\\ 
LogP &  100.0\%$\pm$0.0\% & 0.716$\pm$0.032 &  100.0\%$\pm$0.0\% & 10K+5K\\ 
JNK3+GSK3\bm{$\beta$} & 98.6\%$\pm$1.1\% & 0.721$\pm$0.021 & 91.3\%$\pm$0.6\% & 10K+5K\\ 
QED+SA+JNK3+GSK3\bm{$\beta$} &  99.2\%$\pm$0.3\% & 0.731$\pm$0.029 & 79.4\%$\pm$1.2\% & 20K+5K\\ 
\bottomrule[0.6pt]
\end{tabular}}
\end{table*}

\subsection{Complexity Analysis}
\label{sec:complexity} 

We did computational analysis in terms of oracle calls and computational complexity. 
(1) \textbf{oracle calls}. DST requires $O(TM)$ oracle calls, where $T$ is the number of iterations (Alg~\ref{alg:main}). 
$M$ is the number of generated molecules (Equation.~\ref{eqn:dpp_obj}), we have $M\leq K_{\text{}}J$, $K_{\text{}}$ is the number of nodes in the scaffolding tree, for small molecule, $K_{\text{}}$ is very small. 
$J$ is the number of enumerated candidates in each node. As shown in Figure~\ref{fig:assemble}, $J$ is also upper-bounded ($J\leq 4+14$ for the example in Figure~\ref{fig:assemble}). 
(2) \textbf{computational complexity}. 
The computational complexity is $O(TMC^2)$ (the main bottleneck is DPP method, Algorithm~\ref{alg:dpp_greedy}), where the size of selected molecules $C=10$ for all the tasks (Section~\ref{sec:dpp} \&~\ref{sec:gnn_setup}). 
For all the tasks in Table~\ref{table:denovo_single} and~\ref{table:denovo_multi}, DST can be finished in 12 hours on an Intel Xeon E5-2690 562 machine with 256G RAM and 8 NVIDIA Pascal Titan X GPUs. 
The complexity and runtime are acceptable for molecule optimization.

\section{Additional Experimental Results}
\label{sec:additional_experiment}
In this section, we present the additional empirical results, including additional results on \textit{de novo} generation, ablation study, chemical space visualization, interpretability analysis (case study).

\subsection{Additional results of \textit{de novo} molecular generation}
\label{sec:additional}

In this section, we present some additional results of \textit{de novo} molecular generation for completeness.

\begin{figure}[h!]
\centering
\subfigure[LogP]{
\includegraphics[width=0.32\linewidth]{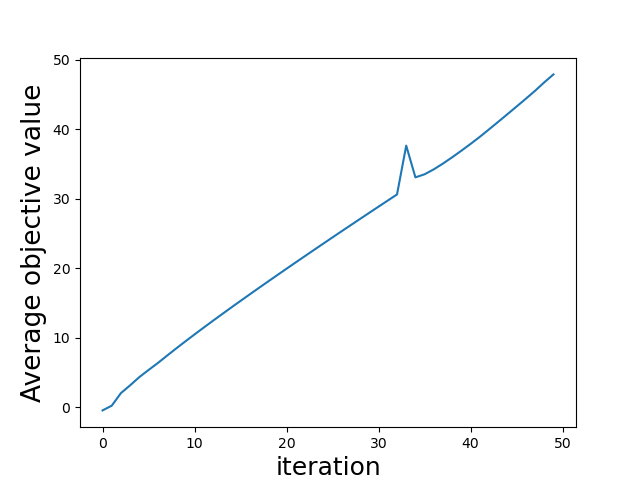}}
\subfigure[JNK3]{
\includegraphics[width=0.32\linewidth]{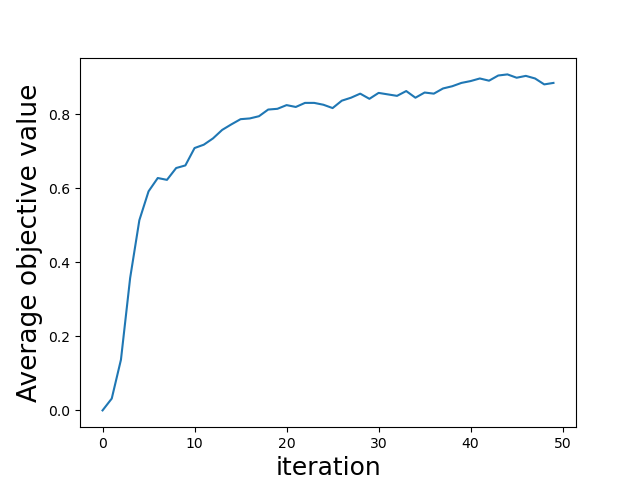}}
\subfigure[GSK3\bm{$\beta$}]{
\includegraphics[width=0.32\linewidth]{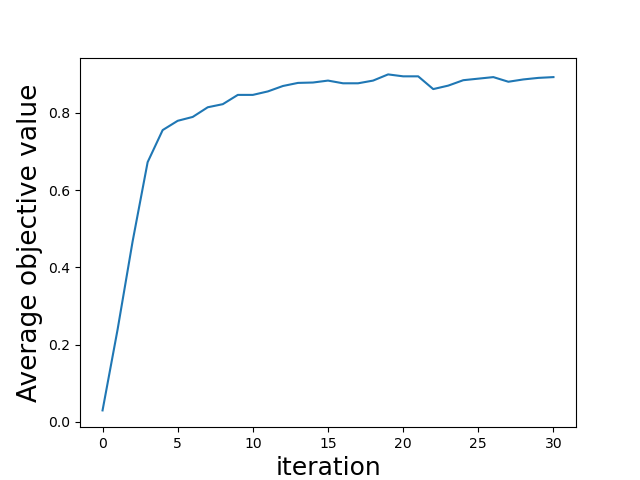}}
\subfigure[QED]{
\includegraphics[width=0.32\linewidth]{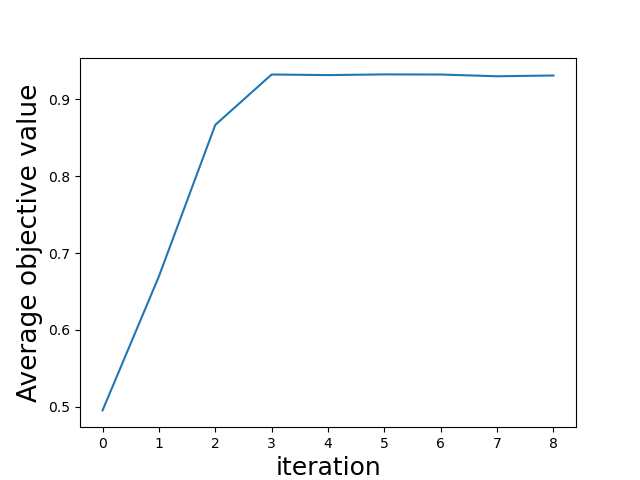}}
\subfigure[JNK3+GSK3\bm{$\beta$}]{
\includegraphics[width=0.32\linewidth]{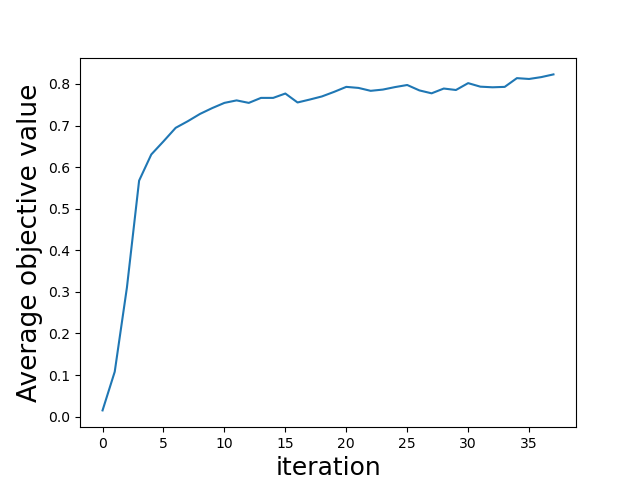}}
\subfigure[QED+SA+JNK3+GSK3\bm{$\beta$}]{
\includegraphics[width=0.32\linewidth]{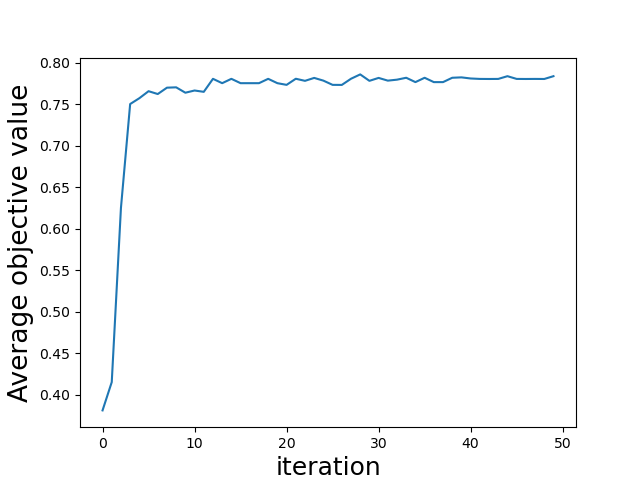}}
\caption{
The optimization curves in \textit{de novo} optimization experiments. The objective value ($F$) is a function of iterations.  }
\label{fig:optimization_curve}
\end{figure}

\begin{figure}[h!]
\centering
\includegraphics[width=\linewidth]{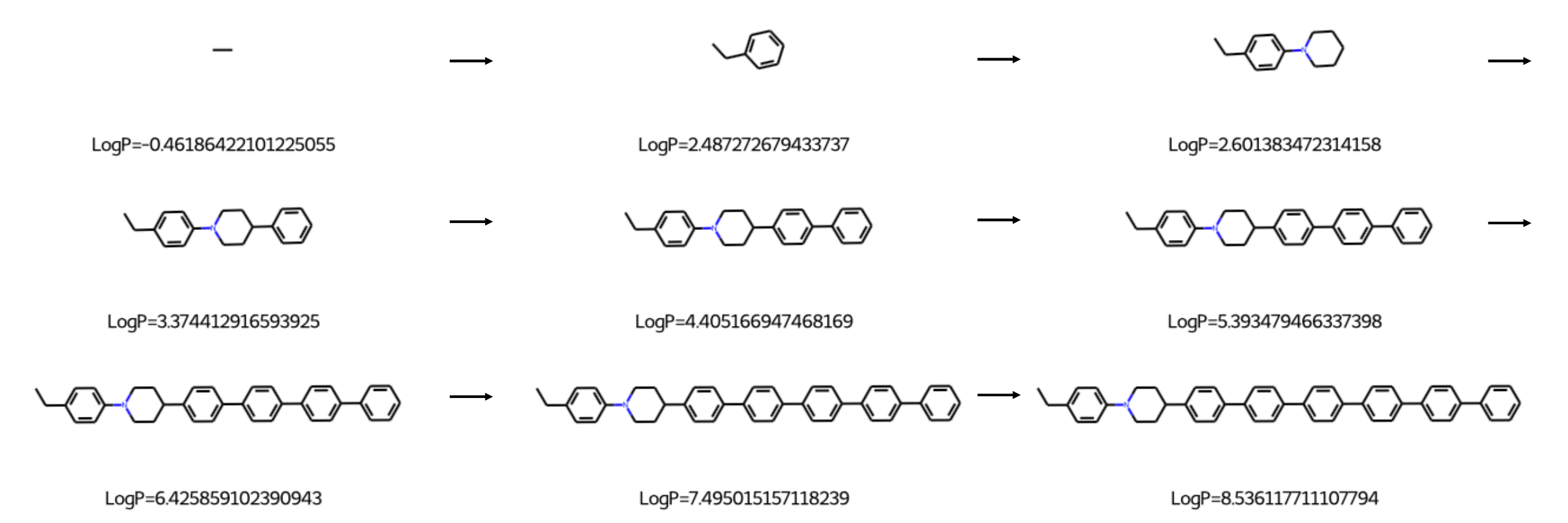}
\caption{
The first eight steps in \textit{the de novo} optimization procedure of LogP. The model successfully learned to add a six-member ring each step. }
\label{fig:logp_growing}
\end{figure}

\begin{figure}[h!]
\centering
\subfigure[Molecules with highest average QED, normalized-SA, JNK3 and GSK3\bm{$\beta$} scores, four scores represent QED, raw SA, JNK3, and GSK3\bm{$\beta$} scores, respectively. ]{\includegraphics[width=13.5cm]{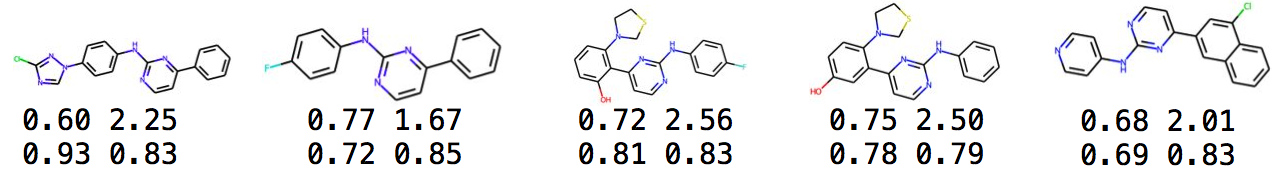}}
\subfigure[Molecules with highest QED. ]{\includegraphics[width=13.5cm]{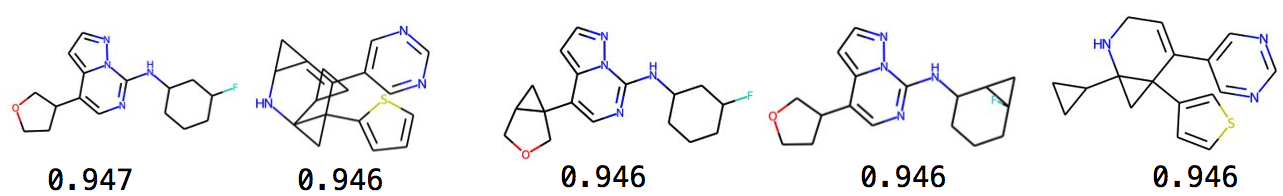}}
\caption{
Sampled molecules with the highest scores. }
\label{fig:highest_example}
\end{figure}

First, we present the optimization curve for all the optimization tasks in Figure~\ref{fig:optimization_curve}. We observe that our method is able to reach a high objective value efficiently within 10 iterations in all the optimization tasks. 
Worth mentioning that when optimizing LogP, the model successfully learned to add a six-member ring each step, as shown in Figure~\ref{fig:logp_growing}, and the objective ($F$) value grows linearly as a function of iteration number, which is theoretically the optimal strategy under our setting. 
Then, in Figure~\ref{fig:highest_example}, we show the molecules with the highest objective ($F$) scores generated by the proposed method on optimizing QED and  ``QED+SA+JNK3+GSK3\bm{$\beta$}''. 
Then we compare our method with baseline methods on 3 molecules with the highest objective ($F$) scores in Table~\ref{table:denovo_top} and~\ref{table:denovo_multi_top} for single-objective and multi-objective generation, respectively.

\begin{table*}[h!]
\tiny
\centering
\caption{Highest scores of generated molecules on single-objective \textit{de novo} molecular generation. We present the result of \mname in the first 50 iterations, but please note the setting of generation varies among the models, and a completely fair comparison is impossible. }
\label{table:denovo_top}
 \resizebox{0.90\textwidth}{!}{
\begin{tabular}{lccc|ccc|ccc}
\toprule[0.6pt]
\multirow{2}{*}{Method} & \multicolumn{3}{c}{JNK3} & \multicolumn{3}{c}{GSK3\bm{$\beta$}} & \multicolumn{3}{c}{LogP} \\
& 1st & 2nd & 3rd & 1st & 2nd & 3rd  & 1st & 2nd & 3rd \\
\hline 
GCPN &  0.57 & 0.56 & 0.54  & 0.57 & 0.56 & 0.56  & 8.0 & 7.9 & 7.8 \\
MolDQN & 0.64 & 0.63 & 0.63 & 0.54 & 0.53 & 0.53  & 11.8 & 11.8 & 11.8  \\ 
GA+D & 0.81 & 0.80 & 0.80 & 0.79 & 0.79 & 0.78 & 20.5 & 20.4 & 20.2 \\
RationaleRL & 0.90 & 0.90 & 0.90  & 0.93 & 0.92 & 0.92 & - & - & - \\
MARS & 0.92 & 0.91 & 0.90 & \bf 0.95 & 0.93 & 0.92  & 45.0 & 44.3 & 43.8 \\
\mname & \bf 0.97 & \bf 0.97 &\bf 0.97 &\bf 0.95 &\bf 0.95 &\bf 0.95 &\bf 49.1 &\bf 49.1 &\bf 49.1 \\ 
\bottomrule[0.6pt]
\end{tabular}}
\end{table*}

\begin{table*}[h!]
\tiny
\centering
\caption{Highest scores of generated molecules on multi-objective \textit{de novo} molecular generation. The score is the average value of all objectives.   }
\label{table:denovo_multi_top}
 \resizebox{0.780\textwidth}{!}{
\begin{tabular}{lccc|ccc}
\toprule[0.6pt]
\multirow{2}{*}{Method} & \multicolumn{3}{c}{JNK3+GSK3\bm{$\beta$}} & \multicolumn{3}{c}{QED+SA+JNK3+GSK3\bm{$\beta$}} \\ 
& top-1 & top-2 & top-3 & top-1 & top-2 & top-3 \\
\hline 
GCPN & 0.31 & 0.31 & 0.30 & 0.57 & 0.56 & 0.56 \\ 
MolDQN & 0.46 & 0.45 & 0.45 & 0.45 & 0.45 & 0.44 \\
GA+D & 0.68 & 0.68 & 0.67 & 0.71 & 0.70 & 0.70\\
RationaleRL & 0.81 & 0.81 & 0.81  & 0.76 & 0.76 & 0.75 \\
MARS & 0.78 & 0.78 & 0.77 & 0.72 & 0.72 & 0.72 \\
\mname & \bf 0.89 & \bf 0.89 & \bf 0.89 & \bf 0.83 & \bf 0.83 & \bf 0.83 \\
\bottomrule[0.6pt]
\end{tabular}}
\end{table*}

\subsection{\textit{De novo} molecular optimization on QED (potential limitation of \DST)}
\label{sec:qed}

\begin{figure}[h!]
\centering
\includegraphics[width=0.45\linewidth]{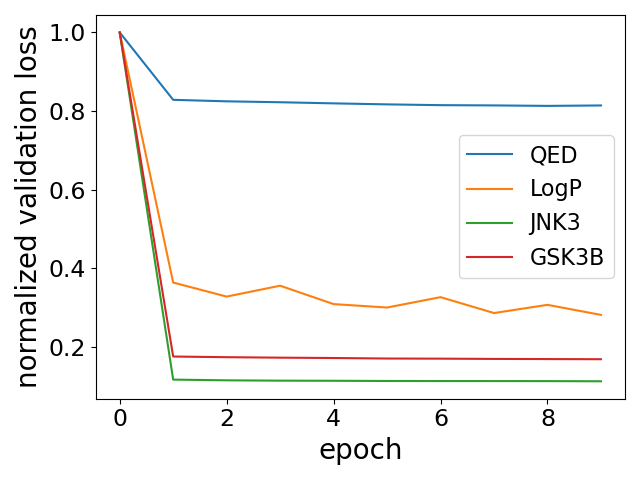}
\caption{Normalized validation loss-epoch learning curves. 
For fairness of comparison, validation loss is normalized by dividing the validation loss at scratch (i.e., 0-th epoch) so that all the validation losses are between 0 and 1. 
For most of the target properties, the normalized loss value on the validation set would decrease significantly, and GNN can learn these properties well, except QED. The key reason for the failure of GNN on optimizing QED is the limitation of the expressive power of scaffolding tree itself. 
QED is a property that is highly dependent on \textit{how} substructures connect~\cite{bickerton2012quantifying}, while our scaffolding tree currently ignores that information. 
See Section~\ref{sec:qed} for more details and analysis.  
}
\label{fig:loss}
\end{figure}

As we have touched in Section~\ref{sec:denovo}, the optimization on QED is not as satisfactory as other oracles. We compare the performance of various methods on single-objective \textit{de novo} molecular generation for optimizing QED score and show the result in Table~\ref{table:denovo_qed}. Additional baseline methods include JTVAE (junction tree variational autoencoder)~\cite{jin2018junction} and GraphAF (Graph Flow-based Autoregressive Model)~\cite{shi2019graphaf}. 
The main reason behind this result is that our GNN predicts the target property based on a scaffolding tree instead of a molecular graph, as shown in Equation~\eqref{eqn:gnn_prediction}. A scaffolding tree omits rings' assembling information, as shown in Figure~\ref{fig:adj}. 
Compared with other properties like LogP, JNK3, GSK3\bm{$\beta$}, drug-likeness is more sensitive to \textit{how} substructures connect~\cite{bickerton2012quantifying}. This behavior impedes the training of GNN and leads to the failure of optimization. 
We report the learning curve in Figure~\ref{fig:loss}, where we plot the normalized loss on the validation set as a function of epoch numbers when learning GNN. 
For fairness of comparison, validation loss is normalized by dividing the validation loss at scratch (i.e., 0-th epoch) so that all the validation losses are between 0 and 1. 
For most of the target properties, the normalized loss value on the validation set would decrease significantly, and GNN can learn these properties well, except QED. 
It verifies the failure of training the GNN on optimizing QED. 
A differentiable molecular graph at atom-wise resolution may potentially solve this problem. 
\begin{table}[h!]
\centering
\caption{Comparison of different methods on optimizing QED for single-objective \textit{de novo} molecular generation. {The results for baseline methods are copied from~\cite{You2018-xh,zhou2019optimization,shi2019graphaf,xie2021mars}. The results of JTVAE are copies from \cite{You2018-xh}}. }
\label{table:denovo_qed}
\resizebox{0.40\textwidth}{!}{
\begin{tabular}{lccc}
\toprule[0.6pt]
Method & top-1 & top-2 & top-3 \\ \hline 
JTVAE~\cite{jin2018junction} & 0.925 & 0.911 & 0.910 \\ 
GCPN~\cite{You2018-xh} & \bf 0.948 & 0.947 & 0.946 \\
MolDQN~\cite{zhou2019optimization} & \bf 0.948 & \bf 0.948 &\bf 0.948  \\
GraphAF~\cite{shi2019graphaf} & \bf 0.948 & \bf 0.948 & 0.947 \\ 
MARS~\cite{xie2021mars} & \bf 0.948  & \bf 0.948 & \bf 0.948 \\
\mname & 0.947 & 0.946 & 0.946 \\ 
\bottomrule[0.6pt]
\end{tabular}}
\end{table}

\subsection{Results Analysis for Distribution Learning Methods (LigGPT)}
As showed in Table~\ref{table:denovo_single} and~\ref{table:denovo_multi}, distribution learning methods (LigGPT)~\cite{bagal2021liggpt} have much weaker optimization ability. 
DST and all the other baselines fall into the category of goal-directed molecule generation, a.k.a., molecule optimization, which generates molecules with high scores for a given oracle. In contrast, LigGPT belongs to distribution learning (a different category of method), which learns the distribution of the training set. We refer to \cite{brown2019guacamol} for more description of two categories of methods.
Consequently, conditioned generation learns from the training set, is unable to generate molecules with property largely beyond the training set distribution and can not optimize a property directly, even though they claim to be able to solve the same problem.  
Problem formulation of distribution learning methods leads to an inability to generate molecules with property largely beyond the training set distribution, which means they are much weaker in optimization.


\subsection{Ablation study}
\label{sec:ablation}
\begin{figure}[h!]
\centering
\subfigure[JNK3]{
\includegraphics[width=0.32\linewidth]{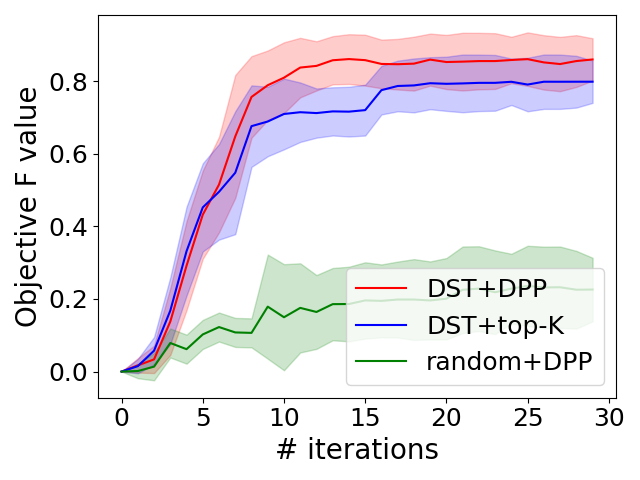}}
\subfigure[GSK3B]{
\includegraphics[width=0.32\linewidth]{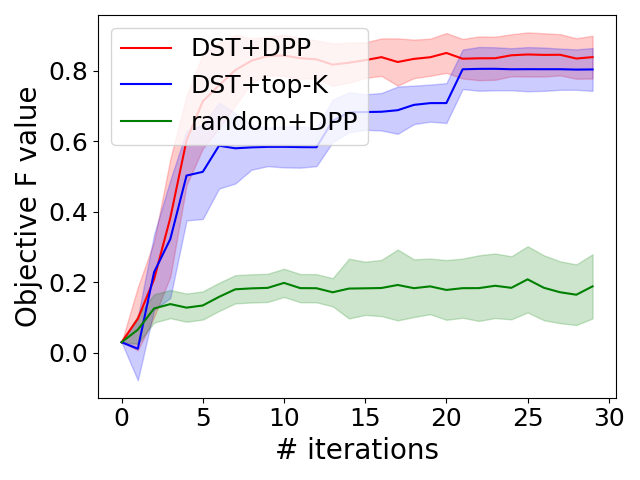}}
\subfigure[LogP]{
\includegraphics[width=0.32\linewidth]{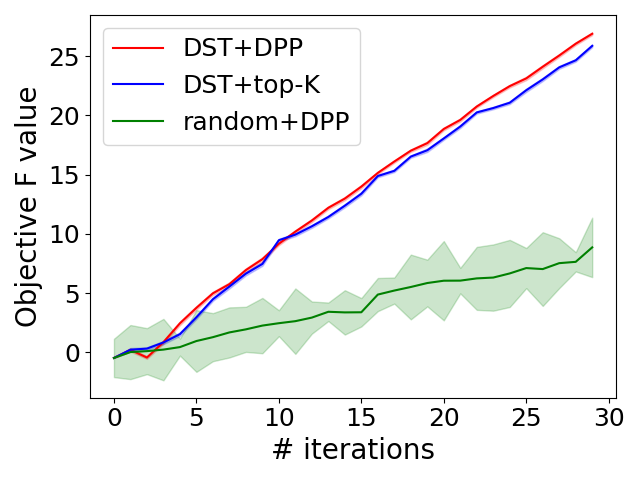}}
\caption{Ablation study. Objective value ($F$) as a function of iterations. See Section~\ref{sec:ablation} for more details. 
}
\label{fig:ablation}
\end{figure}

As described in Section~\ref{sec:differentiable}, during molecule sampling, we sample the new molecule from the differentiable scaffolding tree (Equation~\ref{eqn:sample}). To verify the effectiveness of our strategy, we compare with a \textit{random-walk sampler}, where the topological edition (i.e., expand, shrink or unchange) and substructure are both selected randomly. 
We consider the following variants:
\begin{itemize}
 \item ``\mname + DPP''. Both topology and substructure to fill are sampled from the optimized differentiable scaffolding tree, as shown in Equation~\eqref{eqn:sample}. 
 This is what we use in this paper. 
 \item ``random + DPP''. Changing topology randomly, that is, at each leaf node, ``expand'', ``shrink'' and ``unchange'' probabilities are fixed to $0.5, 0.1, 0.4$. Substructure selection is sampled from the substructures' distribution in the optimized differentiable scaffolding tree. Then it uses DPP (Section~\ref{sec:dpp}) to select diverse and desirable molecules for the next iteration. 
 \item ``\mname + top-$K$''. Same as ``\mname + DPP'', it uses \mname to sample new molecules. The difference is when selecting molecules for the next iteration, it selects the top-$K$ molecules with highest $f$ score. It is equivalent to $\lambda \xrightarrow[]{}+\infty$ in Equation~\eqref{eqn:dpp_obj}. 
\end{itemize}
We show the results in Figure~\ref{fig:ablation}. We find that both \mname sampling and DPP-based diversification play a critical role in performance. 
We check the results for ``\mname + top-K'', during some period, the objective does not grow, we find it is trapped into local minimum, impeding its performance, especially convergence efficiency. 
``random+DPP'' exhibits the random-walk behaviour and it would not reach satisfactory performance. 
When optimizing LogP, ``\mname+DPP'' and ``\mname+top-$K$'' achieved similar performance, because logP score will prefer larger molecules with more carbon atoms, which is less sensitive to the diversity and relatively easier to optimize. Overall, ``\mname+DPP'' is the best strategy compared with other variants. 

\subsection{Chemical space visualization}
\label{sec:visualize_chemical}

\begin{figure}[h!]
\centering
\subfigure[After 3 iterations]{
\includegraphics[width=0.32\linewidth]{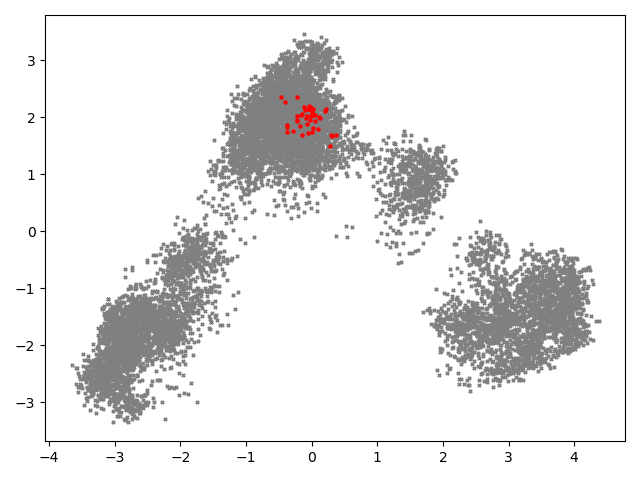}}
\subfigure[After 5 iterations]{
\includegraphics[width=0.32\linewidth]{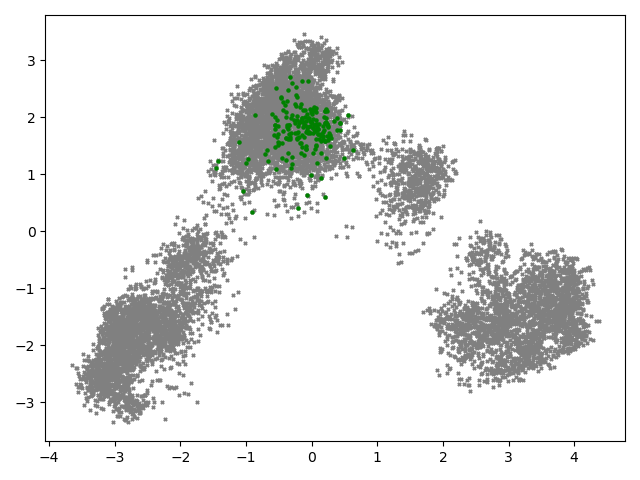}}
\subfigure[After 10 iterations]{
\includegraphics[width=0.32\linewidth]{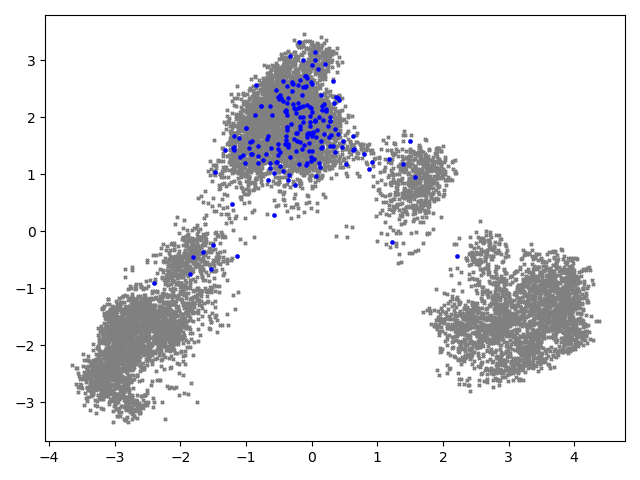}}
\subfigure[After 20 iterations]{
\includegraphics[width=0.32\linewidth]{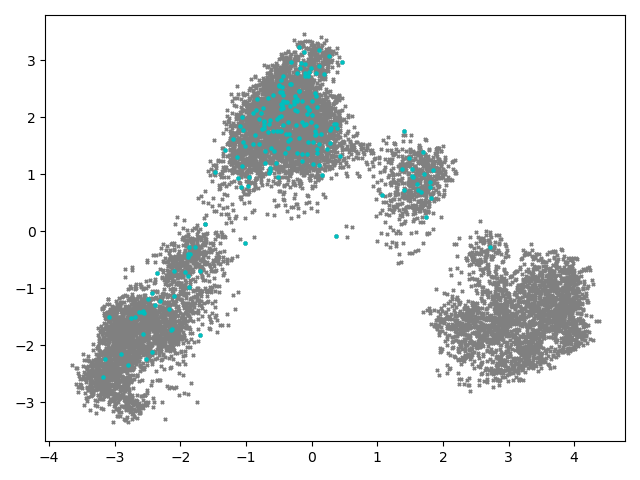}}
\subfigure[After 30 iterations]{
\includegraphics[width=0.32\linewidth]{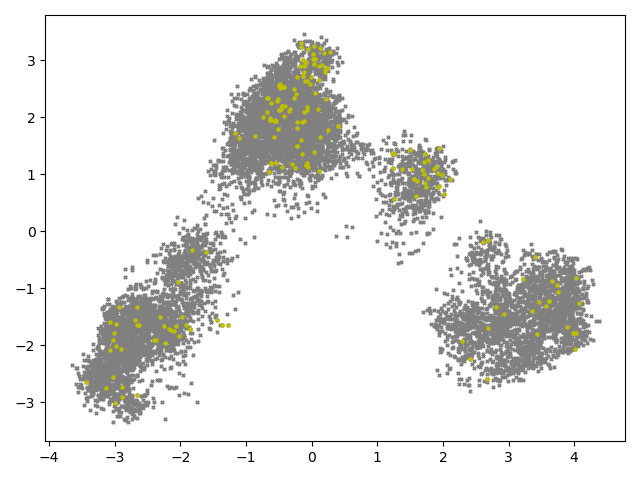}}
\caption{
Visualization of chemical space covered during optimization. We used PCA to reduce the dimension of Morgan fingerprint. The gray points are the ZINC 250k data set. while colored points are generated molecules after corresponding iterations.}
\label{fig:pca}
\end{figure}

We use principle component analysis (PCA) to visualize the distribution of explored chemical structures in optimizing ``JNK3 \& GSK3\bm{$\beta$}''. 
Specifically, we fit a two-dimensional principal component analysis (PCA)~\cite{bro2014principal} for 2048-bit Morgan fingerprint vectors of 20K ZINC molecules, which are randomly selected from ZINC database~\cite{sterling2015zinc}. 
Then we use the PCA to project the fingerprint of the generated molecule from various generations into a two-dimensional vector to observe their trajectories. The results are reported in Figure~\ref{fig:pca}, 
where the grey points represent the two-dimensional vector of ZINC molecules. 
We find that our method explores different parts of the 2D projection of the chemical space and covers a similar chemical space as the ZINC database after 20 to 30 iterations. 

\subsection{Additional Interpretability Analysis}
\label{sec:appendix_interpretability}

\begin{figure}[h!]
\centering
\includegraphics[width=0.6\linewidth]{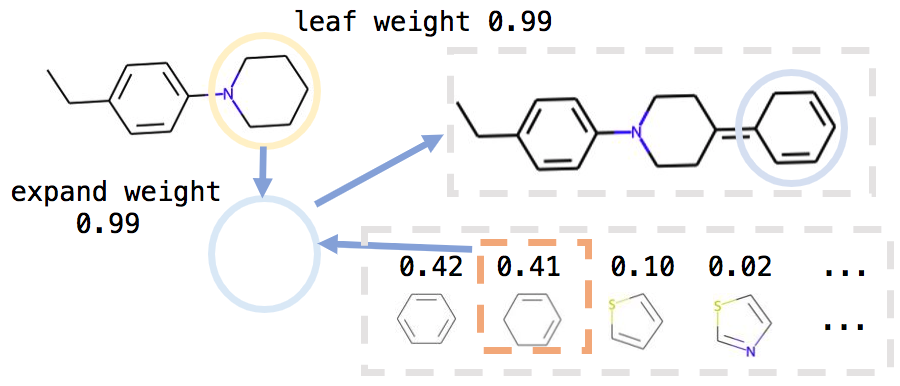}
\caption{Interpretability analysis when optimizing LogP. }
\label{fig:interpret_logp}
\end{figure}

We provide an interpretability example in Figure~\ref{fig:interpret_logp}. At the leaf node (yellow), from the optimized differentiable scaffolding tree, we find that the leaf weight and expand weight are both 0.99. Thus we decide to EXPAND, the six-member ring is selected and filled in the expansion node (blue). This is consistent with our intuition that logP score will prefer larger molecules with more carbon atoms. 

\section{Theoretical Analysis}
\label{sec:additional_theory}
In this section, we present some theoretical results of the proposed method. 
First, in Section~\ref{sec:converge}, we conduct convergence analysis of \mname under certain mild assumptions. 

\subsection{Convergence Analysis}
\label{sec:converge}

In this section, we discuss the theoretical properties of \mname in the context of \textit{de novo} molecule design (learning from scratch). We restrict our attention to a special variant of \mname, named \mname-greedy: at the $t$-th iteration, given one scaffolding tree $Z^{(t)}$, \mname-greedy pick up only one molecule with highest objective value from $Z^{(t)}$'s neighborhood set $\calN(Z^{(t)})$, i.e., $Z^{(t+1)} = {\arg\max}_{Z\in \calN(Z^{(t)})}\ F(Z^{(t)})$ is exactly solved. 
We theoretically guarantee the quality of the solution produced by \mname-greedy. 
First, we make some assumptions and explain why these assumptions hold. 
\begin{assumption}[Molecule Size Bound]
\label{assumption:moleculesize}
The sizes (i.e., number of substructures) of all the scaffolding trees generated by \mname are bound by $N_{\text{min}}$ and $N_{\text{max}}$. 
\end{assumption}
We focus on small molecule optimization; the target molecular properties would decrease greatly when the molecule size is too large, e.g., QED (drug-likeness)~\cite{bickerton2012quantifying}. 
Thus it is reasonable to bound the size of scaffolding tree.  
In addition, we use submodularity and smoothness to characterize the geometry of objective landscape. 
\begin{assumption}[Submodularity and Smoothness]
\label{assumption:submodular}
Suppose $X_{1}, X_{2}, X_{3}$ are generated successively by \mname-greedy via growing (i.e., EXPAND) a substructure on the corresponding scaffolding tree. We assume that the objective gain (i.e., $\Delta F$) brought by adding a single substructure would not increase as the molecule grows (EXPAND). 
\begin{equation}
\begin{aligned}
F(X_3) - F(X_2) \leq F(X_2) - F(X_1), \ \ \ \ \ \  \text{(submodularity)} 
\end{aligned}
\end{equation}
where $X_2 = \text{EXPAND}(X_1, s_1), \ X_3 = \text{EXPAND}(X_2, s_2)$, $s_1, s_2$ are substructures to add. Submodularity plays the role of concavity/convexity in the discrete regime. 
On the other hand, we specify the smoothness of the objective function $F$ by assuming
\[ F(X_3) - F(X_2) \geq \gamma( F(X_2) - F(X_1) ), \ \  0 < \gamma < 1 \ \ \ \  \text{(smoothness)} 
\]
holds for the $X_1,X_2,X_3$ described above, whose sizes are smaller than $N_{\text{min}}$. 
\end{assumption}
Then we theoretically guarantee the quality of the solution under these assumptions. 
\begin{theorem}
\label{theorem:main}
Suppose Assumption~\ref{assumption:moleculesize} and~\ref{assumption:submodular} hold, we have the following relative improvement bound with the optimum 
\begin{equation}
F(Z_*) - F(X_{0}) \geq \frac{1 - \gamma^{N_{\text{min}}}}{(1-\gamma)N_{\text{max}}} \big(F(X_*) - F(X_{0}) \big),
\end{equation}
where $Z_*$ is the local optimum found by \mname-greedy, $X_*$ is the ideal optimal molecule, $X_0$ is an empty molecule, starting point of \textit{de novo} molecule design. 
In molecule generation setting, a molecule is a local optimum when its objective value is maximal within its neighbor molecule set, i.e., $F(Z_*) \geq F(Z)$ for $\forall\ Z \in \calN(Z_*)$. 
\end{theorem}

The proof is given in Section~\ref{sec:theorem}. 

\noindent\textbf{Proof Sketch}. 
We first show that \mname-greedy is able to converge to local optimum within finite step in Lemma~\ref{lemma:local_optimum}. Then we decompose the successive generation path and leverage the geometric information of objective landscape to analyze the quality of local optimum. 
\begin{lemma}[Local optimum]
\label{lemma:local_optimum}
\mname-greedy would converge to local optimum within finite steps. 
\end{lemma} 
The proof is given in Section~\ref{sec:lemma_localoptimum}.


\section{Extension of Molecule diversification}
\label{sec:diversity_theory}

In the current iteration, we have generated $M$ molecules ($X_1, \cdots, X_{M}$) and need to select $C$ molecules for the next iteration. 
We expect these molecules to have desirable chemical properties (high $F$ score) and simultaneously maintain higher structural diversity.

To quantify diversity, we resort to the \textit{determinantal point process (DPP)}~\cite{kulesza2012determinantal}. 
DPP models the repulsive correlation between data points~\cite{kulesza2012determinantal} and has been successfully applied to many applications such as text summarization~\cite{Cho19dpp}, mini-batch sampling~\cite{zhang2017determinantal}, and recommendation system~\cite{chen2018fast}. 
Generally, we have $M$ data points, whose indexes are $\{1,2,\cdots, M\}$, $\bfS \in \RB_{+}^{M\times M}$ denotes the similarity kernel matrix between these data points, $(i,j)$-th element of $\bfS$ measures the Tanimoto similarity between $i$-th and $j$-th molecules. 
We want to sample a subset (denoted $\calR$) of $M$ data,  $\calR$ is a subset of $\{1,2,\cdots, M\}$ with fixed size $C$, it assigns the probability
\begin{equation}
\label{eqn:dpp_appendix}
P(\calR) \propto \det(\bm{S}_{\calR}), \ \ \ \ \text{where}\ \calR\subseteq \{1,2,\cdots, M\},\ |\calR|=C, 
\end{equation}
where $\bm{S}_{\calR} \in \RB^{C\times C}$ is the sub-matrix of $\bfS$, $\det(\bm{S}_{\calR})$ is the determinant of the matrix $\bm{S}_{\calR}$. 
For instance, if we want to sample a subset of size 2, i.e., $\calR = \{i,j\}$, then we have $P(\calR) \propto \det(\bm{S}_{\calR}) = \bm{S}_{ii} \bm{S}_{jj} - \bm{S}_{ij} \bm{S}_{ji} = 1 - \bm{S}_{ij} \bm{S}_{ji}$, more similarity between $i$-th and $j$-th data points lower the probability of their co-occurrence. 
DPP thus naturally diversifies the selected subset. 
DPP can be calculated efficiently using the following method. 
\begin{definition}[DPP-greedy~\cite{chen2018fast}]
\label{def:dppgreedy}
For any symmetric positive semidefinite (PSD) matrix $\bm{S} \in \RB_{+}^{M\times M}$ and fixing the size of $\calR$ to $C$, Problem~\eqref{eqn:dpp_appendix} can be solved in a greedy manner by \textit{DPP-greedy} in polynomial time $O(C^2M)$. 
It is denoted $ {\calR} = \text{DPP-greedy}(\{X_1,\cdots,X_M\}, C)$. 
\end{definition}
We describe the DPP-greedy algorithm in Algorithm~\ref{alg:dpp_greedy} for completeness. During each iteration, it selects one data sample that maximizes the current objective, as described in Step 5 in Algorithm~\ref{alg:dpp_greedy}.

\begin{algorithm}[h!]
\caption{DPP-greedy~\cite{chen2018fast}} 
\label{alg:dpp_greedy}
\begin{algorithmic}[1]
\STATE \textbf{Input}: symmetric positive semi-definite matrix $\bfS \in \RB^{M\times M}$, number of selected data $C \in \mathbb{N}_{+}$, $C < M$. 
\STATE \textbf{Output}: $\calR \subseteq\{1,2,\cdots, M\}, |\calR|=C$.  
\STATE $\calW = \{1,2,\cdots,M\}$. 
\FOR{$i=1,2,\cdots,C$}
\STATE $j = \underset{k\in \calW}{\arg\max}\ \ \log \det(S_{\calR \cup \{k\}})$. 
\STATE $\calR = \calR \cup \{j\}$. 
\STATE $\calW = \calW - \{j\}$. 
\ENDFOR 
\end{algorithmic}
\end{algorithm}

As mentioned, our whole target is to select the molecules with desirable properties while maintaining the diversity between molecules. 
The objective is formulated as 
\begin{equation}
\label{eqn:dpp_obj_appendix}
\underset{\calR \subseteq\{1,2,\cdots, M\}, |\calR|=C}{\arg\max}\ \calL_{\text{DPP}}(\calR) = \lambda \sum_{t\in \calR} v_t + \log P(\bm{S}_{\calR}) = \log \det(\bm{V}_{\calR}) + \log \det(\bm{S}_{\calR}), 
\end{equation}
where the hyperparamter $\lambda > 0$ balances the two terms, the diagonal matrix $\bm{V}$ is 
\begin{equation}
\label{eqn:v_appendix}
\bm{V} = \text{diag}\big([\exp(\lambda v_{1}), \cdots, \exp(\lambda v_{M})] \big), \ \ \ \ \ \text{where}\ v_1 = F(X^{}_1), \cdots, v_{M} = F(X^{}_{M}), 
\end{equation}
where $v_i$ is the $F$-score of the $i$-th molecule (Eq.~\ref{eqn:objective}), 
$\bm{V}_{\calR}$ is a sub-matrix of $\bfV$ indexed by $\calR$.  
For any square matrix $\bm{M}_1, \bm{M}_2$ of the same shape, we have
\[
\det(\bm{M}_1\bm{M}_2) = \det(\bm{M}_1) \det(\bm{M}_2) = \det(\bm{M}_2) \det(\bm{M}_1),
\]
 we further transform $\calL_{\text{DPP}}(\calR)$ as  below to construct symmetric matrix,
\begin{equation}
\label{eqn:dpp_obj2_appendix}
\calL_{\text{DPP}}(\calR) = \log \det(\bm{V}_{\calR}) + \log \det(\bm{S}_{\calR}) = \log \det(\bm{V}_{\calR} \bm{S}_{\calR}) =  \log \det \Big( \bm{V}_{\calR}^{\frac{1}{2}} \bm{S}_{\calR} \bm{V}_{\calR}^{\frac{1}{2}} \Big), 
\end{equation}
where 
$\bm{V}^{\frac{1}{2}} = \text{diag} \big( [\exp(\frac{\lambda v_{1}}{2}), \cdots,  \exp(\frac{\lambda v_{M}}{2})] \big)$. 
Then we present the following lemma for the usage of the \textit{DPP-greedy} method. 
\begin{lemma}
\label{lemma:psd}
Suppose $\bfS \in \RB^{M\times M}$ is the (Tanimoto) similarity kernal matrix of the $M$ molecules, i.e., $\bm{S}_{ij} = \frac{\bfb_i^\top \bfb_j}{\Vert \bfb_i\Vert_2 \Vert \bfb_j\Vert_2}$, $\bfb_i$ is the binary fingerprint vector for the $i$-th molecule, $V$ is diagonal matrix defined in Eq.~\eqref{eqn:v_appendix}, 
then we have (1) $\bm{V}^{\frac{1}{2}} \bfS \bm{V}^{\frac{1}{2}}$ is positive semidefinite; \ (2) $\bm{V}_{\calR}^{\frac{1}{2}} \bm{S}_{\calR} \bm{V}_{\calR}^{\frac{1}{2}} = (\bm{V}^{\frac{1}{2}} \bfS \bm{V}^{\frac{1}{2}})_{\calR}$. 
\end{lemma}
The proof is given in Section~\ref{sec:lemma_psd}.

Thus, Problem~\eqref{eqn:dpp_obj2_appendix} can be transformed as 
\begin{equation}
\label{eqn:dpp_obj3}
\underset{\calR \subseteq\{1,2,\cdots, M\}, |\calR|=C}{\arg\max}\ \calL_{\text{DPP}}(\calR) =  \log \det \Big( \big(\bm{V}^{\frac{1}{2}} \bfS \bm{V}^{\frac{1}{2}}\big)_{\calR} \Big),  
\end{equation}
which means we can use \textit{DPP-greedy} (Def.~\ref{def:dppgreedy}) to solve Problem~\eqref{eqn:dpp_obj2_appendix} and obtain the optimal $\calR$.

\textbf{Discussion}. 
In Eq.~\eqref{eqn:dpp_obj_appendix}, we have two terms to specify the constraints on molecular property and structural diversity, respectively. 
When we only consider the first term ($\lambda \sum_{t\in \calR} v_t$), the selection strategy is to select $C$ molecules with the highest $F$ score for the next iteration, same as conventional evolutionary learning in~\cite{brown2019guacamol,jensen2019graph,nigam2019augmenting}.

On the other hand, if we only consider the second term in Eq.~\eqref{eqn:dpp_obj_appendix}, we show the effect of selection strategies under certain approximations. 
Suppose we have $C$ molecules $X_1,X_2,\cdots,X_{C}$ with high diversity among them, then we leverage \mname to optimize these $C$ molecules respectively, and obtain $C$ clusters of new molecules, i.e., $\hat{Z}_{11},\cdots, \hat{Z}_{1l_1} \overset{\text{i.i.d.}}{\sim}  \text{DMG-Sampler}(\widetilde{\bfN}^{*}_{({X_1})}, \widetilde{\bfA}^{*}_{(X_1)}, \widetilde{\bfw}^{*}_{(X_1)}); \cdots; \hat{Z}_{{C}1},\cdots, \hat{Z}_{{C}l_{C}} \overset{\text{i.i.d.}}{\sim}  \text{DMG-Sampler}(\widetilde{\bfN}^{*}_{({X_{C}})}, \widetilde{\bfA}^{*}_{(X_{C})}, \widetilde{\bfw}^{*}_{(X_{C})})$. 
Then we present the following lemma to show that when only considering diversity, under certain assumptions, Problem~\eqref{eqn:dpp_obj_appendix} reduces to multiple chain MCMC methods.


In Eq.~\eqref{eqn:dpp_obj_appendix}, $\lambda$ is a key hyperparamter, a larger $\lambda$ corresponds to more weights on objective function $F$ while smaller $\lambda$ specifies more diversity. 
When $\lambda$ goes to infinity, i.e., only considering the first term ($\lambda \sum_{t\in \calR} v_t$), it is equivalent to selecting $C$ molecule candidates with the highest $F$ score for the next iteration, same as conventional evolutionary learning in~\cite{jensen2019graph,nigam2019augmenting}.

On the other hand, if we only consider the second term, we show the effect of selection strategies under certain approximations. 
Suppose we have $C$ molecules $X_1,X_2,\cdots,X_{C}$ with high diversity among them, then we leverage \mname to optimize these $C$ molecules respectively, and obtain $C$ clusters of new molecules, i.e., 
\begin{equation*}
\begin{aligned}
& \hat{Z}_{11},\cdots, \hat{Z}_{1l_1} \overset{\text{i.i.d.}}{\sim}  \text{DST-Sampler}(\widetilde{\bfN}^{*}_{({X_1})}, \widetilde{\bfA}^{*}_{(X_1)}, \widetilde{\bfw}^{*}_{(X_1)}); \\
& \cdots; \\
& \hat{Z}_{{C}1},\cdots, \hat{Z}_{{C}l_{C}} \overset{\text{i.i.d.}}{\sim}  \text{DST-Sampler}(\widetilde{\bfN}^{*}_{({X_{C}})}, \widetilde{\bfA}^{*}_{(X_{C})}, \widetilde{\bfw}^{*}_{(X_{C})})
\end{aligned}
\end{equation*}
Then we present the following lemma to show that when only considering diversity, under certain assumptions, Problem~\eqref{eqn:dpp_obj_appendix} reduces to multiple independent Markov chain. 
\begin{lemma}
\label{lemma:approx}
Assume (1) the inter-cluster similarity is upper-bounded, i.e., $\text{sim}(\hat{Z}_{ip}, \hat{Z}_{jq})\leq \epsilon_1$ for any $i\neq j$; (2) the intra-cluster similarity is lower-bounded, i.e., $\text{sim}(\hat{Z}_{ip}, \hat{Z}_{iq})\geq 1 - \epsilon_2$ for any $i \in \{1,2,\cdots, M\}$ and $p\neq q$; 
when both $\epsilon_1, \epsilon_2$ approach to $0_+$, the optimal solution to Problem~\eqref{eqn:dpp_obj_appendix} is 
\[
\{\hat{Z}_{1p_1}, \hat{Z}_{2p_2}, \cdots, \hat{Z}_{Cp_{C}}\}, 
\]
{where} $p_{c} = {\arg\max}_{p} F(\hat{Z}_{cp})\ \text{for}\ c=1,\cdots,C$. 
\end{lemma}
The proof is given in Section~\ref{sec:lemma_diversity}. 

\noindent\textbf{Remark}. When the inter-cluster similarity is low enough, and intra-cluster similarity is high enough, our molecule selection strategy reduces to multiple independent Markov chains. However, these assumptions are usually too restrictive for small molecules.

\section{Proof of Theoretical Results}
\label{sec:proof}
In this section, we provide the proof of all the theoretical results in Section~\ref{sec:additional_theory} and~\ref{sec:diversity_theory}. 

\subsection{Proof of Lemma~\ref{lemma:psd}}
\label{sec:lemma_psd}

\begin{proof}

(I) $\bfV^{\frac{1}{2}} \bfS \bfV^{\frac{1}{2}}$ is positive semidefinite.

First, let us prove similarity kernel matrix $\bfS \in RB^{M\times M}$ based on molecular Tanimoto similarity is positive semidefinite (PSD), we know that the $(i,j)$-th element of $\bfS$ measures the Tanimoto similarity between $i$-th and $j$-th molecules, i.e., 
\[
\bfS_{ij} = \frac{\bfb_i^\top \bfb_j}{\Vert \bfb_i\Vert_2 \Vert \bfb_j\Vert_2}, 
\]
where $\bfb_i \in [0,1]^{P}$ is the $P$-bit fingerprint vector for the $i$-th molecule (in this paper, $P=2048$). 
$\bfS$ can be decomposed as 
\begin{equation*}
\bfS = \bfB \bfB^\top,
\end{equation*}
where matrix $\bfB$ is the stack of all the normalized (divided by $l_2$ norm, $\Vert\cdot\Vert_2$) fingerprint vector, as 
\begin{equation*}
\bfB = \bigg[ \frac{\bfb_1}{\Vert \bfb_1\Vert_2}, \frac{\bfb_2}{\Vert \bfb_2\Vert_2}, \cdots, \frac{\bfb_{P}}{\Vert \bfb_{P}\Vert_2} \bigg] \in \RB^{P\times M}. 
\end{equation*}
For $\forall\ \bfx \in \RB^{M}$, we have 
\begin{equation*} 
\bfx^\top \bfS \bfx = \bfx^\top \bfB^\top \bfB \bfx = (\bfB \bfx)^\top (\bfB \bfx) \geq 0. 
\end{equation*}
Thus, $\bfS$ is PSD. 

Then, similarly, for $\forall\ \bfx \in \RB^{M}$, we have 
\begin{equation*}
\bfx^\top \bfV^{\frac{1}{2}} \bfS \bfV^{\frac{1}{2}} \bfx = \bfx^\top \big( \bfV^{\frac{1}{2}}\big)^\top \bfB^\top \bfB \bfV^{\frac{1}{2}} \bfx = (\bfB \bfV^{\frac{1}{2}} \bfx)^\top (\bfB \bfV^{\frac{1}{2}} \bfx) \geq 0. 
\end{equation*}
where $\bfV^{\frac{1}{2}}$ is diagonal matrix, so $\bfV^{\frac{1}{2}} = (\bfV^{\frac{1}{2}})^\top$. 
Thus, $\bfV^{\frac{1}{2}} \bfS \bfV^{\frac{1}{2}}$ is symmetric and positive semidefinite. 

(II) $\bfV_{\calR}^{\frac{1}{2}} \bfS_{\calR} \bfV_{\calR}^{\frac{1}{2}} = (\bfV^{\frac{1}{2}} \bfS \bfV^{\frac{1}{2}})_{\calR}$. 

Without loss of generalization, we assume $\calR = \{t_1, \cdots, t_C\}$, where $t_1 < t_2 < \cdots, t_C$. $\bfV^{\frac{1}{2}}$ is diagonal. 
\begin{equation*}
\bfV_{\calR}^{\frac{1}{2}} = \begin{bmatrix}\exp( \frac{\lambda v_{t_1}}{2}  ) & & \\ & \ddots & \\ & & \exp( \frac{\lambda v_{{t_C}}}{2} )\end{bmatrix}, 
\end{equation*}
where 
\[
v_{t_i} = F(X_{t_i})
\]
is the objective function of $t_i$-th molecule $X_{t_i}$. 
The $i,j$-th element of $\bfV_{\calR}^{\frac{1}{2}} \bfS_{\calR} \bfV_{\calR}^{\frac{1}{2}}$ is 
\begin{equation}
\label{eqn:vsv_a}
\bigg( \bfV_{\calR}^{\frac{1}{2}} \bfS_{\calR} \bfV_{\calR}^{\frac{1}{2}}\bigg)_{ij} =  \exp\big(\frac{\lambda v_{t_i}}{2}\big) \bfS_{t_it_j} \exp\big( \frac{\lambda v_{t_j}}{2} \big). 
\end{equation}

On the other hand, the $i,j$-th element of $\bfV^{\frac{1}{2}} \bfS \bfV^{\frac{1}{2}}$ is $\exp(\frac{\lambda v_{i}}{2}) \bfS_{ij} \exp(\frac{\lambda v_{j}}{2})$. 
Then the $i,j$-th element of $\bigg(\bfV^{\frac{1}{2}} \bfS \bfV^{\frac{1}{2}}\bigg)_{\calR}$ is
\begin{equation}
\label{eqn:vsv_b}
\Big(\big(\bfV^{\frac{1}{2}} \bfS \bfV^{\frac{1}{2}}\big)_{\calR} \Big)_{ij} = \exp\big(\frac{\lambda v_{t_i}}{2}\big) \bfS_{t_it_j} \exp\big(\frac{\lambda v_{t_j}}{2} \big). 
\end{equation}

Combining Equation~\eqref{eqn:vsv_a} and~\eqref{eqn:vsv_b}, we prove  $\bfV_{\calR}^{\frac{1}{2}} \bfS_{\calR} \bfV_{\calR}^{\frac{1}{2}} = (\bfV^{\frac{1}{2}} \bfS \bfV^{\frac{1}{2}})_{\calR}$. 

\end{proof}

\subsection{Proof of Lemma~\ref{lemma:approx}}
\label{sec:lemma_diversity}
\begin{proof}

We consider two cases in the solution $\calR$. (A) one molecule for each input molecule $Z_1, \cdots, Z_C$. (B) other cases. Our solution belongs to Case (A). 

(A) First, we prove for (A), our solution is optimal. 
We consider the second term in Equation~\eqref{eqn:dpp_obj_appendix}, $\bfS_{\calR}$ is diagonal dominant. 
Also, determinant function is a continuous function with regard to all the elements. 
Thus, $det(\bfS_{\calR}) = \prod_{i=1}^{C} (\bfS_{\calR})_{ii}$ goes to 1.  
Intuitively, all the selected molecules are dissimilar to each other and the diversity is maximized. 
On the other hand, to maximizing the first term in Equation~\eqref{eqn:dpp_obj_appendix}, during each $k\in \{1,2,\cdots,C\}$, we select molecule with highest $F$ score from $\{\hat{Z}_{k1},\cdots, \hat{Z}_{kl_{k}}\}$. That is our solution. 

(B) Then we prove all the possible combinations in (B) are worse than our solution. 
In (B), based on pigeonhole principle, there are at least one input molecule $Z_k$ that corresponds to at least two selected molecules. Without loss of generalization, we denoted them $\hat{Z}_{k_1}$ and $\hat{Z}_{k_2}$. Since $\bfS_{\calR}$ is diagonal dominant, its determinant can be decomposed as 
\[
\det(\bfS_{\calR}) = \prod_{k=1}^{C} \det(\hat{S}_k).  
\]
If there is at least one $\hat{S}_k$ whose shape is greater than 1. 
Based on definition of determinant, for matrix $A\in \RB^{M\times M}$
\begin{equation}
\det(\bfA) = \sum_{\eta \in \text{Perm}(M)} \text{sgn}(\eta) \prod_{i=1}^{M} \bfA_{i,\eta(i)},
\end{equation}
where $\text{Perm}(M)$ is the set of all permutations of the set $\{1,2,\cdots,M\}$, 
$\text{sgn}(\eta)$ denotes the signature of $\eta$, a value that is +1 whenever the reordering given by $\eta$ can be achieved by successively interchanging two entries an even number of times, and -1 whenever it can be achieved by an odd number of such interchanges. 
For exactly half of all $\eta$s, $\text{sgn}(\eta) = 1$ and the other half are equal to -1. For the matrix $A$ whose shape is greater than 1 and all the elements are equal to 1, the determinant is equal to 
$\sum_{\eta \in \text{Perm}(M)} \text{sgn}(\eta) = 0$.

Determinant function is a continuous function with regard to all the elements. 
When $\epsilon_2$ goes to $0_{+}$, all the elements of $\hat{S}_k$ approach to 1, the determinant goes to 0. 
Thus, $\det(\bfS_{\calR})$ also goes to 0. 
The objective in Equation~\eqref{eqn:dpp_obj_appendix} goes to negative infinity. 
Thus, it is worse than our solution. Proved. 

\end{proof}

\subsection{Proof of Lemma~\ref{lemma:local_optimum}}
\label{sec:lemma_localoptimum}
\begin{proof}
For the \textit{de novo} design, \mname-greedy start from scratch (empty molecule). 
First, we show in this setting, there is no ``REPLACE'' or ``DELETE'' by mathematical induction and contradiction. 
Since we start from an empty molecule, at the 1-st step the action is ``EXPAND''. 
Then we show the first $t$ steps are ``EXPAND'', the $(t+1)$-th step is still ``EXPAND''. 
Now we have $X^{(t)} = \text{EXPAND}(X^{(t-1)}, s_{t-1})$ (where $s_{t-1}$ is a substructure).
Suppose the $(t+1)$-th step \mname's action is ``REPLACE'', e.g., $X^{(t+1)} = \text{REPLACE}(X^{(t)}, s_t)$ (where $s_t$ is a substructure), based on definition of \mname-greedy, we have $F(X^{(t+1)}) \geq F(X^{(t)})$. Since \mname only REPLACE the leaf node, we find that $X^{(t+1)}$ and $X^{(t)}$ are both in neighbor molecule set of $X^{(t-1)}$, i.e., $\calN(X^{(t-1)})$, which contradict with the fact that $X^{(t)} = {\arg\max}_{X\in \calN(X^{(t-1)})}\ F(X^{(t-1)})$. Similarly, we show that there would not exist ``DELETE''. 

Then based on Assummption~\ref{assumption:moleculesize}, we find that \mname-greedy converges at most $N_{\text{max}}$ steps. 

\end{proof}

\subsection{Proof of Theorem~\ref{theorem:main}}
\label{sec:theorem}
\begin{proof}

Based on the Proof of Lemma~\ref{lemma:local_optimum}, we find that there is only ``EXPAND'' action, then we are able to decompose the generation path as follows. 
Starting from scratch, i.e., $X_0$, suppose the path to optimum $X_{*}$ is 
\[
X_0 \xrightarrow[]{} X_1 \xrightarrow[]{} X_2 \xrightarrow[]{} \cdots \xrightarrow[]{} X_{k_1} = X_*,
\]
where each step one substructure is added. 
The path produced by \mname-greedy is 
\[
Z_0 (X_0) \xrightarrow[]{} Z_1 \xrightarrow[]{} Z_2 \xrightarrow[]{} \cdots \xrightarrow[]{} Z_{k_2} = Z_*.
\]
Based on the definition of  in each step only one substructure is added.  

Based on Assumption~\ref{assumption:moleculesize}, we have $N_{\text{min}} \leq k_1, k_2 \leq N_{\text{max}}$. 
There might be some overlap within the first several steps, without loss of generalization, we assume $Z_k = X_k$ and $Z_{k+1} \neq X_{k+1}$, where $k$ can be $0, 1, \cdots, k_1$. 
Based on Assumption~\ref{assumption:submodular}, we have 
\begin{equation*}
F(X_1) - F(X_0) \geq F(X_2) - F(X_1) \geq F(X_3) - F(X_2) \geq \cdots  \geq F(X_{k_1}) - F(X_{k_1 - 1}). 
\end{equation*}
Then, we have 
\begin{equation*}
\begin{aligned}
& k_1 \big(F(X_1) - F(X_0)\big)
\\ 
\geq & \big(F(X_1) - F(X_0)\big) + \big( F(X_2) - F(X_1)\big) + \big( F(X_3) - F(X_2)\big) + \cdots + \big(F(X_{k_1}) - F(X_{k_1 - 1}) \big) \\
= & F(X_{k_1}) - F(X_{0}) \\ 
\end{aligned}
\end{equation*}
Thus, we get
\begin{equation}
\label{eqn:x1_x0}
\begin{aligned}
& F(X_1) - F(X_0) \geq  \frac{1}{k_1} \big( F(X_{k_1}) - F(X_0) \big) \\
\geq &  \frac{1}{N_{\text{max}}} \big( F(X_{k_1}) - F(X_0) \big) = \frac{1}{N_{\text{max}}} \big( F(X_*) - F(X_0) \big). 
\end{aligned}
\end{equation}
Since $Z_0 = X_0$, according to the definition of greedy algorithm, we have $F(Z_1) \geq F(X_1)$. 
Based on Assumption~\ref{assumption:submodular}, we have 
\begin{equation*}
\begin{aligned}
F(Z_{N_{\text{min}}}) - F(Z_{N_{\text{min}} - 1}) 
& \geq \gamma \big(F(Z_{N_{\text{min}}-1}) - F(Z_{N_{\text{min}}-2}) \big) \geq \gamma^2 \big( F(Z_{N_{\text{min}}-2}) - F(Z_{N_{\text{min}}-3}) \big) \\ 
& \geq \cdots \geq \gamma^{N_{\text{min}} - 1} \big(F(Z_1) - F(Z_0) \big). 
\end{aligned}
\end{equation*}
Based on Assumption~\ref{assumption:moleculesize}, we have $ F(Z_*) - F(Z_0) \geq F(Z_{N_{\text{min}}}) - F(Z_0) $. 
Then we have 
\begin{equation}
\label{eqn:zstar_z0}
\begin{aligned}
& F(Z_*) - F(Z_0) \\
\geq &  F(Z_{N_{\text{min}}}) - F(Z_0)  \\
=  & \big( F(Z_{N_{\text{min}}}) - F(Z_{N_{\text{min}}-1}) \big) + \big( F(Z_{N_{\text{min}}-1}) - F(Z_{N_{\text{min}}-2}) \big) + \cdots + \big( F(Z_1) - F(Z_0)\big)  \\ 
\geq & (1 + \gamma + \gamma^2 + \cdots \gamma^{N_{\text{min}}-1}) \big(F(Z_1) - F(Z_0) \big) \\ 
= & \frac{1 - \gamma^{N_{\text{min}}}}{1-\gamma} \big( F(Z_1) - F(Z_0)\big) 
\end{aligned}
\end{equation}
Combining Equation~\eqref{eqn:x1_x0} and \eqref{eqn:zstar_z0}, we have 
\begin{equation*}
F(Z_*) - F(X_0) \geq \frac{1 - \gamma^{N_{\text{min}}}}{(1-\gamma)N_{\text{max}}} \big(F(X_*) - F(X_0) \big). 
\end{equation*}
We observe that objective $F$'s improvement is relatively lower bounded. 

\end{proof}





\end{document}